\documentclass[lettersize,journal]{IEEEtran}
\usepackage{amsmath,amsfonts}
\usepackage{algorithmic}
\usepackage{algorithm}
\usepackage{array}
\usepackage[caption=false,font=normalsize,labelfont=sf,textfont=sf]{subfig}
\usepackage{textcomp}
\usepackage{stfloats}
\usepackage{url}
\usepackage{verbatim}
\usepackage{graphicx}
\usepackage{cite}
\usepackage {subcaption}
\usepackage{makecell}

\usepackage{booktabs}
\usepackage{multirow}
\usepackage{makecell}
\usepackage{xspace}
\usepackage{threeparttable}
\begin{document}

\title{Towards Open-Set Myoelectric Gesture Recognition via Dual-Perspective Inconsistency Learning}

\author{Chen Liu$^{\dag}$, Can Han$^{\dag}$, Chengfeng Zhou, Crystal Cai, and Dahong Qian
\thanks{$\dag$These authors contributed to the work equally and should be regarded as co-first authors. (Corresponding author: Dahong Qian.)}
\thanks{This work was partially supported by OYMotion Technologies.}
\thanks{The authors are with the School of Biomedical Engineering, Shanghai Jiao Tong University, Shanghai 200240, China. (email: lchen1206@sjtu.edu.cn; hancan@sjtu.edu.cn; chengfengzhou@sjtu.edu.cn; crystal.cai@sjtu.edu.cn; dahong.qian@sjtu.edu.cn).}
}


\maketitle

\begin{abstract}
Gesture recognition based on surface electromyography (sEMG) has achieved significant progress in human-machine interaction (HMI), especially in prosthetic control and movement rehabilitation. However, accurately recognizing predefined gestures within a closed set is still inadequate in practice; a robust open-set system needs to effectively reject unknown gestures while correctly classifying known ones, which is rarely explored in the field of myoelectric gesture recognition. To handle this challenge, we first report a significant distinction in prediction inconsistency discovered for unknown classes, which arises from different perspectives and can substantially enhance open-set recognition performance. Based on this insight, we propose a novel dual-perspective inconsistency learning approach, PredIN, to explicitly magnify the prediction inconsistency by enhancing the inconsistency of class feature distribution within different perspectives. Specifically, PredIN maximizes the class feature distribution inconsistency among the dual perspectives to enhance their differences. Meanwhile, it optimizes inter-class separability within an individual perspective to maintain individual performance. Comprehensive experiments on various benchmark datasets demonstrate that the PredIN outperforms state-of-the-art methods by a clear margin. Our proposed method simultaneously achieves accurate closed-set classification for predefined gestures and effective rejection for unknown gestures, exhibiting its efficacy and superiority in open-set gesture recognition based on sEMG.
\end{abstract}

\begin{IEEEkeywords}
open-set recognition, gesture recognition, dual-perspective, inconsistency learning, surface electromyography.
\end{IEEEkeywords}

\section{Introduction}
\begin{figure}[!t]
    \vspace{-0.5cm}
    \hspace{-2cm}
    \centering
    \includegraphics[width=1.0\columnwidth]{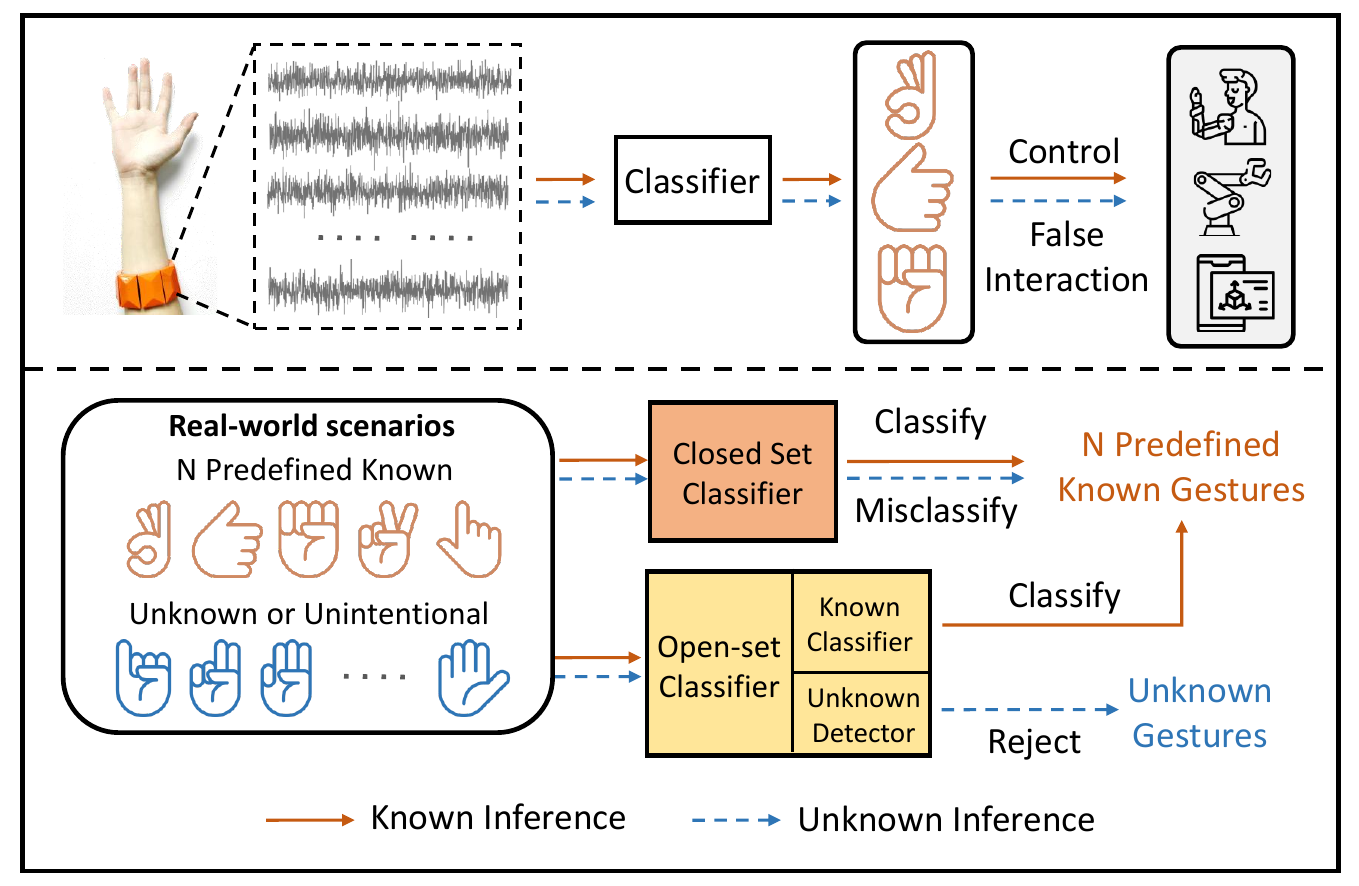}
    \hspace{-2cm}
    \caption{\textbf{An illustration of open-set myoelectric gesture recognition system.} An open-set myoelectric gesture recognition system need to correctly classify predefined known gestures while rejecting unknown gestures to avoid generating false interaction signals.}
\label{fig0}
\end{figure}

\IEEEPARstart{I}{n} the human-machine interaction (HMI) paradigm, gesture recognition serves as a foundational task and has been extensively applied across diverse domains, such as movement rehabilitation~\cite{dere2023event}, prosthetic control~\cite{tang2024exosuit} and mobile interaction~\cite{kang2022reduce}. Recently, the development of gesture recognition systems~\cite{islam2022application, islam2024surface, xiong2024patchemg} based on surface electromyography (sEMG) signals has been remarkable. However, most of them are confined to classic closed-set scenarios, where the training and test sets share an identical label space. These closed-set systems lack robustness and reliability in the dynamic and ever-changing real world, which causes them to mistake novel gestures or unintentional motions as known ones and generate false interaction signals. Therefore, a robust gesture recognition system, one which can correctly classify predefined known gestures while identifying unknown gestures in real-world scenarios (\figurename~\ref{fig0}), is in high demand. Scheirer et al.~\cite{Scheirer2013Open} first described the above demand as open-set recognition (OSR), whose test set contains unknown classes that are not included in the training set.

OSR is an active topic in the field of computer vision, with numerous methods continuously being proposed. However, only a few studies~\cite{wu2021metric, Wu2022UnknownMR} focus on open-set myoelectric gesture recognition. Due to the inherently random and non-stationary nature of sEMG signals, commonly used methods based on reconstruction or generative models in OSR may not be applicable, particularly in achieving closed-set classification accuracies comparable to discriminative methods~\cite{furui2021emg, huang2022class}.
A predominant aspect of existing OSR discriminative methods is to explore the distinctions between known and unknown classes, and design various strategies to enlarge them~\cite{park2024understanding}. Accordingly, a score function is derived based on these distinctions to reject the unknown. A recently popular trend for OSR is employing prototype learning (PL) since it establishes a clear distance distinction between the known and unknown, and demonstrates more promising performance than softmax prediction probabilities distinction~\cite{Yang2020ConvolutionalPN, Chen2022ARPL}. PL methods are able to learn a compact feature space while keeping open space for the unknown. Although the methods based on PL have achieved promising performance, they do not fully explore the inherent distinctions between known and unknown classes as distinguishing them from a single perspective is insufficient.

\begin{figure*}[!t]
    \centering
    \hspace{-1.0cm}
    \begin{minipage}[b]{0.44\textwidth}
        \centering
	  \includegraphics[width=\textwidth]{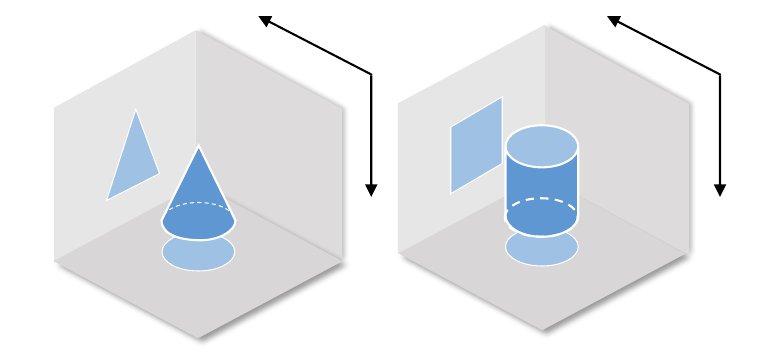}
        \label{fig1:c}
        \caption*{(a) A toy example}
    \end{minipage}
    \begin{minipage}[b]{0.26\textwidth}
        \centering
        \includegraphics[width=\textwidth]{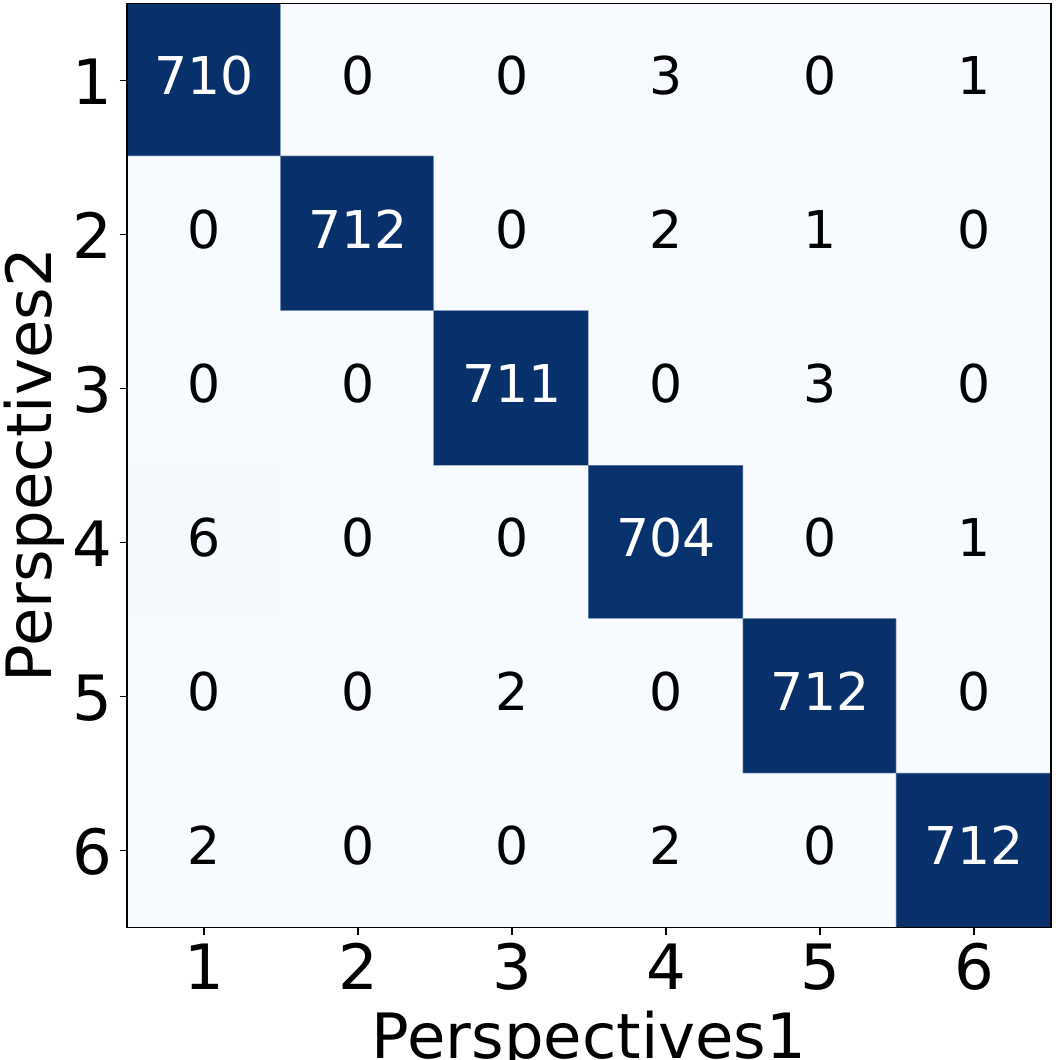}
        \label{fig1:a}
        \vspace{-0.5cm}
        \caption*{(b) Known samples}
    \end{minipage}
     \hspace{0.2cm}
    \begin{minipage}[b]{0.26\textwidth}
        \centering
	  \includegraphics[width=\textwidth]{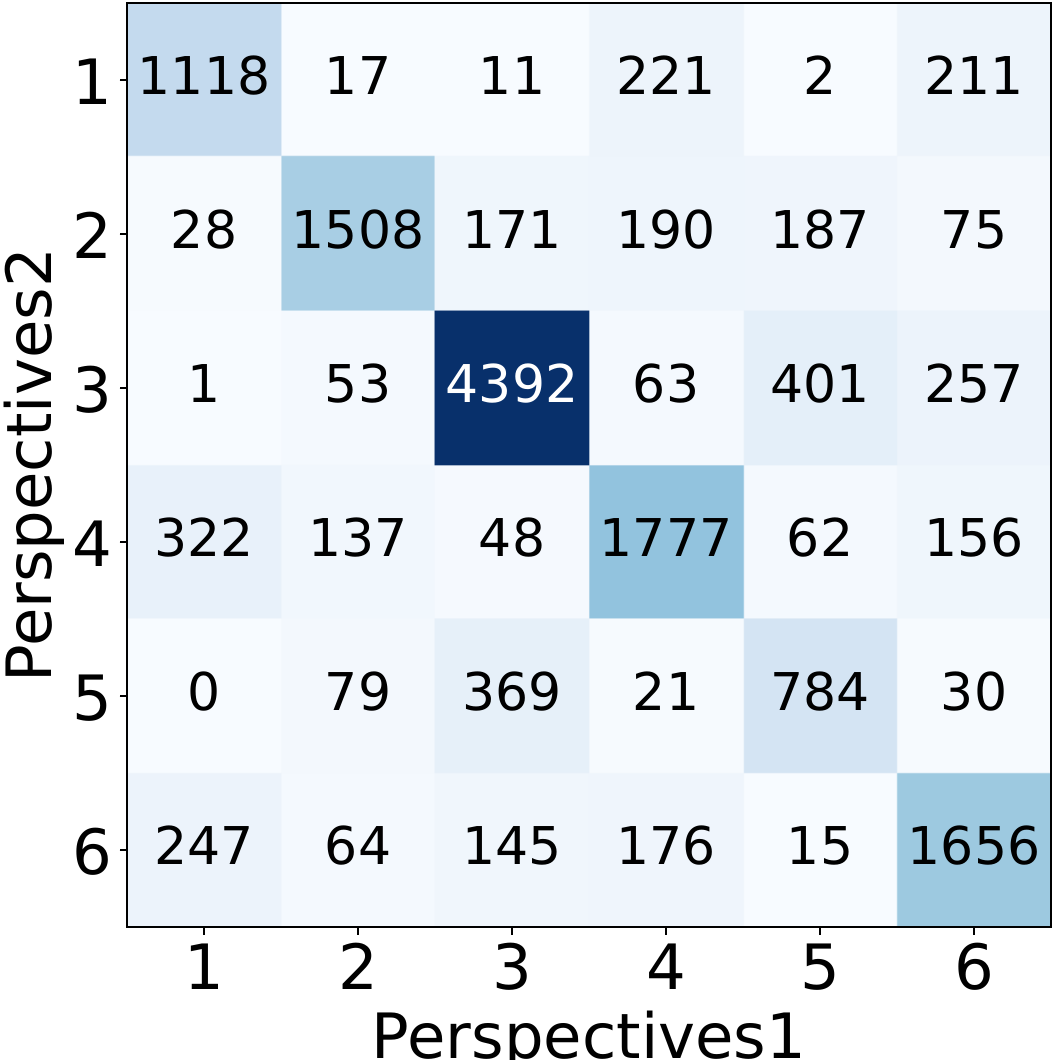}
        \label{fig1:b}
        \vspace{-0.5cm}
        \caption*{(c) Unknown samples}
    \end{minipage}
    \hspace{-1.2cm}
    \vspace{-0.1cm}
    \caption{\textbf{An illustration of prediction inconsistency for the unknown between different perspectives.} We summarize the prediction results of samples in the two perspectives and present them as confusion matrices. Each value in the matrices represents the number of known or unknown samples classified into certain known classes within two networks. Both horizontal and vertical coordinates represent class labels. There are pronounced distinctions in prediction inconsistency between known and unknown samples. (a) is a toy example about different perspectives. (b) and (c) represent the prediction results of bioDB2 samples.}
    \vspace{-0.4cm}
    \label{fig1}
\end{figure*}

Beyond the distinction in distance, we reveal that prediction inconsistency within different perspectives can boost the OSR performance. In real-world scenarios, a single entity can be described in various aspects depending on the perspectives. By considering multiple perspectives, a more reliable assessment of the true similarity between two entities can be made. Unexpectedly, these diverse perspectives play a crucial role in identifying the unknown according to our findings. A toy example is shown in \figurename~\ref{fig1}(a), we could represent a known sample as a cone and an unknown sample as a cylinder. Imagine projecting the cone and cylinder: the projection of the cylinder appears pseudo-similar to the cone when viewed from the top, but different from another angle. Nonetheless, the projection of the cone retains its resemblance to other cones, regardless of the projection direction. Similarly, a known sample will be consistent with other known samples of the same class from any projection direction, constituting prediction inconsistency (\figurename~\ref{fig1}(b)). However, it is difficult for an unknown sample to maintain the pseudo-similarity with the same known class across different perspectives, leading to its prediction inconsistency (\figurename~\ref{fig1}(c)). To better understand the prediction inconsistency, it is important to note that classification models will assign improperly high confidence for unknown samples and misclassify them into known classes~\cite{Yang2020ConvolutionalPN}. The distinction in prediction inconsistency facilitates the differentiation of unknown samples. In our example, the two perspectives, represented by two identical networks, differ solely due to the randomness of the initialization and the learning procedure~\cite{rame2021dice}. Despite this, it is promising that it exhibits pronounced distinctions in prediction inconsistency between known and unknown samples. Learning to incorporate significantly diverse perspectives will help magnify the prediction inconsistency for the unknown. In light of this, a natural idea is to enhance the differences between perspectives. 

To this end, we propose a novel dual-perspective inconsistency learning approach, PredIN, to magnify the prediction inconsistency by explicitly enhancing the inconsistency of class feature distribution within perspectives. Specifically, PredIN constructs two branches representing dual perspectives. Among the two perspectives, PredIN maximizes the inconsistency of class feature distribution by inconsistency loss to enhance perspective differences. Within an individual perspective, PredIN incorporates a triplet loss to optimize inter-class separability, thereby maintaining individual performance. These two strategies act complementarily to regularize the class feature distribution. PredIN ultimately rejects the unknown based on prediction inconsistency and distance. We conduct comprehensive experiments on public datasets to validate the superiority of our proposed method. 

{\bf Summary of Contributions}
\begin{itemize}
    \item [1)] A novel dual-perspective inconsistency learning framework, PredIN, is designed to address a crucial and underexplored task in open-set myoelectric gesture recognition. We reveal a significant distinction between known and unknown samples in open-set scenarios, where prediction inconsistency within different perspectives can notably boost open-set recognition performance. 
    \item [2)] To magnify the distinction in prediction inconsistency, we propose two complementary strategies that learn significantly diverse perspectives by explicitly maximizing the class feature distribution inconsistency while maintaining individual performance.
    \item [3)] Comprehensive experiments on multiple public sEMG datasets demonstrate that our approach simultaneously maintains closed-set classification accuracy for known gestures and improves rejection for unknown gestures, outperforming previous approaches by a clear margin. 
\end{itemize}

\section{Related works}
\label{sec2}

\subsection{Closed-Set sEMG-based Gesture Recognition}

The emergence of deep learning has freed sEMG-based gesture recognition from the constraints of manual feature extraction, facilitating a better understanding of human gestures~\cite{Xiong2021Review}. Various deep learning architectures have been widely employed for this task. Park et al.~\cite{Park2016MovementID} pioneered the application of Convolutional Neural Network (CNN) models to classify the Ninapro DB2 dataset~\cite{DB2}. Furthermore, more complex CNN models and Recurrent Neural Network (RNN) models have showcased their superiority to fine gesture classification~\cite{Xiong2021Review}. EMGHandNet~\cite{karnam2022emghandnet} proposed a hybrid CNN and Bi-LSTM framework to capture both the inter-channel and temporal features of sEMG. 
The attention mechanism is also popular in this field, due to the natural electrode channels and spatial attributes of sEMG signals. Liu et al.~\cite{liu2024transformer} proposed a Transformer-based network to capture global contextual information and local details for sparse sEMG recognition.

Although these progresses have been made, closed-set gesture recognition systems are fragile and may generate false interaction signals when facing inferences from novel gestures or intentional muscle contractions, leading to reduced system reliability and user experience. These extraordinary performances of classic closed-set systems are inadequate since their applications are limited when it comes to the real and open world. In contrast, our work aims to develop an open-set gesture recognition system which can correctly classify known gestures while rejecting unknown gestures in real-world scenarios.

\subsection{Open-Set Recognition}

Open-set recognition seeks to generalize the recognition tasks from a closed-world assumption to an open set. The main challenge that exists in the OSR tasks is the semantic shift where the labels in the training set and testing set are different~\cite{sun2023OSRsurvey}. Existing methods can be mainly divided into discriminative methods which learn rejection rules directly and generative methods which model the distribution of known or unknown classes~\cite{sun2023OSRsurvey}.

Previous OSR discriminative methods established the rejection rules or distinctions mainly in prediction probability~\cite{Bendale2016Openmax, Zhou2021Palceholder} and distance~\cite{Cevikalp2023From, sun2024overall}. Bendale and Boult~\cite{Bendale2016Openmax} demonstrated the limitation of softmax probabilities and introduced OpenMax, a new model prediction layer based on extreme value theory. CPN~\cite{Yang2020ConvolutionalPN} was the first to introduce prototype learning to OSR, modeling known classes as prototypes and rejecting the unknown based on distance metric. ARPL~\cite{Chen2022ARPL} considered the potential characteristics of the unknown data and proposed the concept of reciprocal points to introduce unknown information. Subsequently, more methods based on prototype learning have been proposed, focusing on improving the compactness of known features~\cite{xia2023slcpl}, mining high-quality and diverse prototypes~\cite{Lu2022PMALOS}, or constructing multiple Gaussian prototypes for each class~\cite{liu2023MGPL}. In addition to distance metric, Park et al.~\cite{park2024understanding} observed the distinction in the Jacobian norm between the known and unknown and devised an m-OvR loss to induce strong inter-class separation within the known classes. MEDAF~\cite{wang2024exploring} proposed that learning diverse representations is the key factor in OSR. Beyond the discriminative methods, numerous researchers believe that modeling only known classes is insufficient and suggest incorporating prior knowledge about unknown classes by generative models. Some approaches attempted to generate fake data~\cite{moon2022difficulty}, counterfactual images~\cite{Neal2018OSCI} or confused samples~\cite{Chen2022ARPL}. 

Despite advancing OSR performance in image recognition, only a few studies~\cite{wu2021metric, Wu2022UnknownMR} have focused on the challenge of open-set myoelectric gesture recognition. Wu et al.~\cite{wu2021metric} identified the unknown based on distinctions in distance and reconstruction error through metric learning and autoencoders (AE). To avoid the high computational complexity of generative models, Wu et al.~\cite{Wu2022UnknownMR} further introduced the convolutional prototype network (CPN) to construct multiple prototypes for known classes, employing a matching-based approach to reject the unknown. While these methods have made progress, there still needs to be further exploration to enhance the performance of open-set sEMG-based gesture recognition.

Different from the above methods, our approach emphasizes the distinction in prediction inconsistency and distance metric to reject the unknown. In light of these distinctions, we propose a discriminative approach based on the dual-perspective inconsistency learning framework, which encourages maximizing differences between perspectives. In this paper, we demonstrate that prediction inconsistency within the dual-perspective inconsistency learning framework shows superiority in unknown rejection.

\section{METHODOLOGY}
\label{sec3}

\subsection{Problem Definition}

\begin{figure*}[!t]
    \vspace{-0.5cm}
    \hspace{-2cm}
    \centering
    \includegraphics[width=\textwidth]{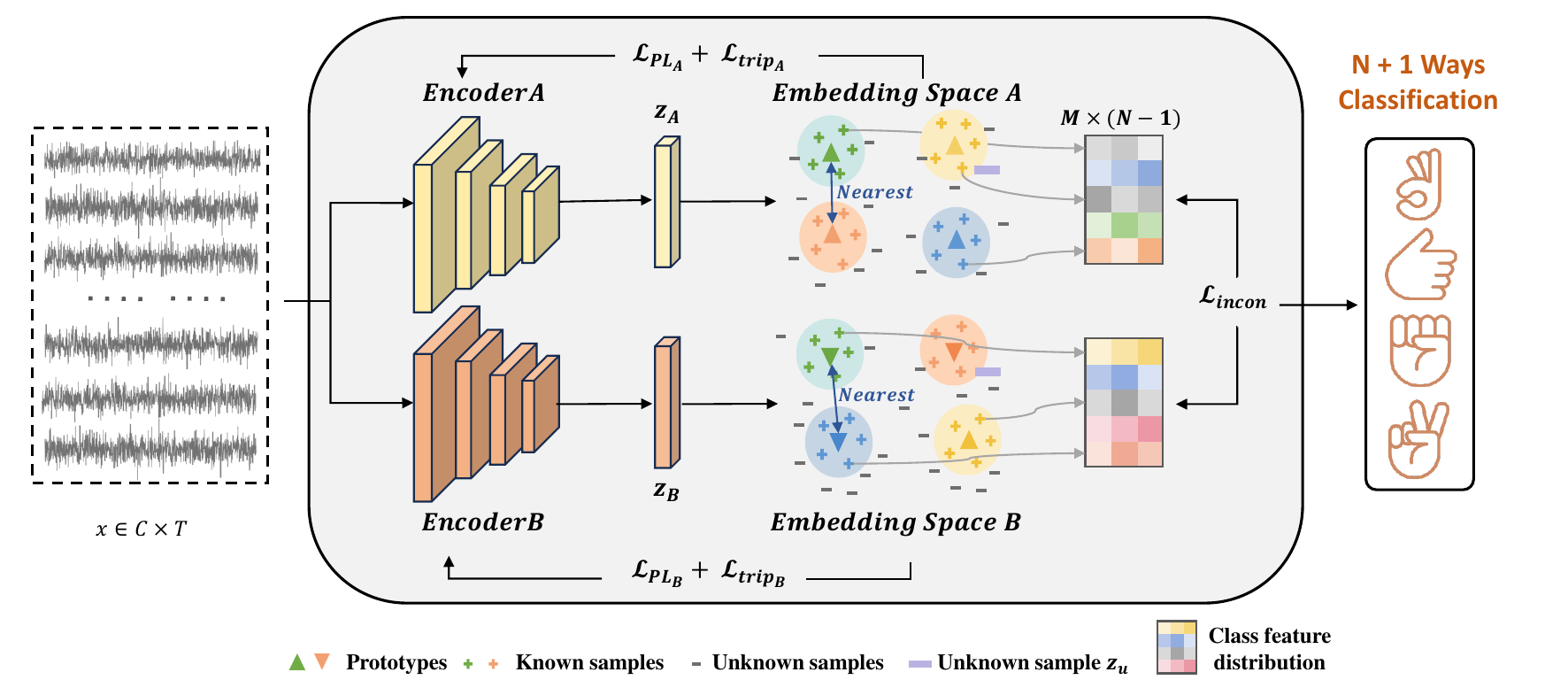}
    \hspace{-2cm}
    \caption{\textbf{An illustration of our proposed framework.} Our framework consists of dual perspectives represented by two branches, each of which contains an encoder and a set of learnable prototypes. $\mathcal{L}_{PL}$ and $\mathcal{L}_{trip}$ are applied to each branch individually while $\mathcal{L}_{incon}$ simultaneously acts on both. $\mathcal{L}_{incon}$ aims to maximize the class feature distribution inconsistency, ensuring each branch has an entirely distinct layout or neighboring class pairs (blue arrows). Upon this, unknown samples like $\mathbf{z}_{u}$ represented by purple are predicted near the clusters of different known classes due to prediction inconsistency while known samples agree on the same predictions across two perspectives.}
    \vspace{-0.4cm}
\label{fig2}
\end{figure*}

Considering the sEMG-based gesture recognition in real-world scenarios, we assume that $ \mathcal{Y} \subset \mathbb{N} $ is the infinite label space of all possible gesture classes. Assume that $\mathcal{C} = \{1,\ldots,N\} \subset \mathcal{Y}$ represents $N$ known classes of interest, which appear in both the training and test sets. The set $\mathcal{U} = \mathcal{Y} \backslash \mathcal{C}$ represents all unknown classes that need to be rejected and only appear in the test set. The objective of open-set recognition is to find a measurable recognition function $f^*\subset \mathbb{H}$ which minimizes both the empirical classification risk on known samples and the open space risk on unknown samples. Open space risk refers to the risk of incorrectly labeling any unknown class as a known one~\cite{Scheirer2013Open}.

\begin{equation}
f^*={ \underset{f}{\arg\min}} \{R_\epsilon(f,D_c) + R_O(f, D_u)\}
\label{eq1}
\end{equation}
where $D_c$ and $D_u$ represent samples belonging to known and unknown classes, respectively.

\subsection{Methodology Overview}
To minimize both the empirical classification risk and open space risk simultaneously, we propose a novel dual-perspective inconsistency learning approach, PredIN. Our proposed framework is illustrated in \figurename~\ref{fig2}. Specifically, we simultaneously train two branches representing the two perspectives. Each branch contains an encoder $h$ with an arbitrary architecture and $N$ learnable prototypes of known classes. Prototype learning acts as a basic classification module, with the loss $\mathcal{L}_{PL}$ applied to each branch to learn a clear distance distinction. Apart from serving as a classification module, PL naturally aligns with proposed two strategies of regularizing the class feature distribution (detailed in Section~\ref{sec3.4} and Section~\ref{sec3.5}). Specifically, to enlarge the distinction in prediction inconsistency, the inconsistency loss $\mathcal{L}_{incon}$ operates concurrently on both branches to enhance their differences by maximizing the inconsistency of class feature distribution among perspectives. In addition, we apply the loss $\mathcal{L}_{trip}$ to enhance inter-class separability, thereby maintaining the individual performance.

\subsection{Prototype Learning}
Each individual perspective of PredIN employs prototype learning to establish a clear distance distinction. The core idea of PL methods is to encourage samples to be close to their corresponding prototypes and distant from others, where class prototypes serve as centers or representatives of each class~\cite{Yang2020ConvolutionalPN}. This establishes a compact feature space and a closed classification boundary for known classes while preserving open space for unknown samples~\cite{Chen2022ARPL}. It provides a distance-based approach for rejecting unknown samples, which has been proven superior to softmax-based approaches~\cite{Lu2022PMALOS}.

For a given sample $\mathbf{x}_i \in \mathbb{R}^{C\times T}$ ($C$ denotes the number of sEMG channels and $T$ denotes the number of timesteps) with label $y_i$, its embedding feature is defined as $\mathbf{z}_i = h(\mathbf{x}_i) \in \mathbb{R}^d$. $N$ known classes are each assigned a learnable prototype $\mathbf{p}^k \in \mathbb{R}^d$, where $1 \leq k \leq N$. The probability of the prediction result $\hat{y_i}$ being $k$ for $\mathbf{x}_i$ is based on the distance $d(h(\mathbf{x}_i),\mathbf{p}^k)$:
\begin{equation}
p(\hat{y_i}=k | \mathbf{x}_i, h, \mathbf{p}) = \frac{e^{-d(h(\mathbf{x}_i),\mathbf{p}^k)}}{\sum_{j=1}^N e^{-d(h(\mathbf{x}_i), \mathbf{p}^j)}}.
\label{eq2}
\end{equation}

To narrow the distance between samples and their corresponding prototypes while pushing them away from other prototypes, the DCE loss function~\cite{Yang2020ConvolutionalPN} is utilized and described as follows:
\begin{equation}
\mathcal{L}_\epsilon = -\frac{1}{M}\sum_{i=1}^M \log{(p(\hat{y_i}=k|\mathbf{x}_i,h,\mathbf{p}))},
\label{eq3}
\end{equation}
where $M$ represents the number of known samples.
We use the dot product to measure the generalized distance between the samples and prototypes. 

The DCE loss only guarantees the discriminability of feature space. To enhance the intra-class compactness, we incorporate an additional compactness term as
\begin{equation}
\mathcal{L}_{com} = \frac{1}{M}\sum_{i=1}^M \mathcal{L}_n(h(\mathbf{x}_i)-\mathbf{p}^k),
\label{eq5}
\end{equation}
in which
\begin{equation}
\mathcal{L}_n(\mathbf{u}) = 
\begin{cases}
\frac{1}{2} \Vert \mathbf{u} \Vert_2 &\quad \Vert \mathbf{u} \Vert_1 < 1\\
\Vert \mathbf{u} \Vert_1 -\frac{1}{2} &\quad \Vert \mathbf{u} \Vert_1 \geq 1 \\
\end{cases},
\label{eq6}
\end{equation}
where $y_i = k$.

Combining \eqref{eq3} and \eqref{eq5}, the overall loss function $\mathcal{L}_{PL}$ of the PL baseline is expressed as follows:
\begin{equation}
\mathcal{L}_{PL} = \mathcal{L}_\epsilon+\beta \mathcal{L}_{com} ,
\label{eq7}
\end{equation}
where $\beta$ controls the intensity of $\mathcal{L}_{com}$.

\subsection{Dual-perspective Class Feature Distribution Inconsistency}
\label{sec3.4}

\begin{figure}[!t]
    \centering
        \includegraphics[width=\columnwidth]{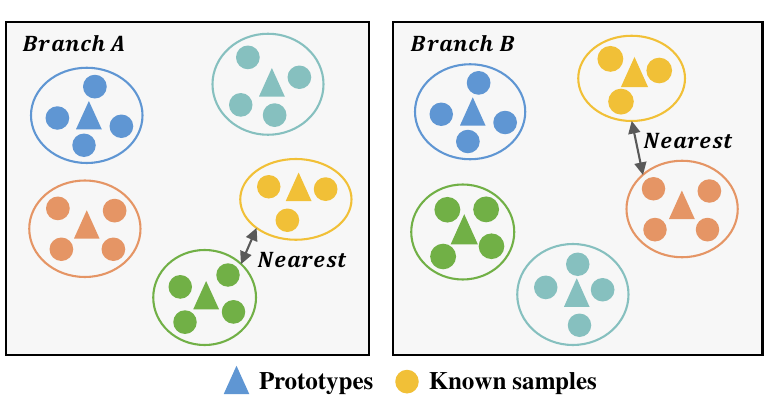}
        \label{featuredistribution} 
    \vspace{-0.5cm}
    \caption{\textbf{An illustration of class feature distribution inconsistency between dual perspectives.} Different colors represent different classes.}
    \vspace{-0.3cm}
    \label{fig3}
\end{figure}

In deep learning, class feature distribution is derived from the projection of high-dimensional sample space to low-dimensional feature space with an encoder. Given this, maximizing class feature distribution inconsistency among different perspectives can promote diverse mapping functions, thereby enhancing their differences. Globally, class feature distribution inconsistency refers to the entirely distinct layout or relative positions of deep features from different classes. Locally, it means that the neighboring classes for each class among branches differ, as illustrated in \figurename~\ref{fig3}.

Based on PL, the class feature distribution can be approximately characterized by learned prototypes since the actual distribution is not directly accessible and prototypes provide the first-order statistics of the distribution~\cite{wen2019comprehensive}. We compute the distance between each sample feature $\mathbf{z_i}$ and prototypes $\mathbf{p
}^j$ of non-corresponding classes from the same branch:
\begin{equation}
d(\mathbf{z}_i,\mathbf{p}^j) = -{\mathbf{z}_i}^T \cdot {\mathbf{p}^j},
\label{eq8}
\end{equation}
where $1 \leq j \leq N$ and $j \neq y_i$.

To compare class feature distribution across different feature spaces, we convert distances into probabilities that signify proximity. In this procedure, we have two criteria: firstly, the closer the distances between features, the higher the corresponding probability values. Secondly, adjusting the relative positions between two widely separated classes has a relatively minor impact on the change of class feature distribution compared to adjusting local structures. Therefore, we aim to focus more on neighboring classes. Softmax meets these two requirements, which adjusts the global layout but highlights the influence of neighboring classes: 
\begin{equation}
p(\mathbf{z}_i,\mathbf{p}^k) = \frac{e^{-d(\mathbf{z}_i,\mathbf{p}^k)}}{\sum_{j=1}^N e^{-d(\mathbf{z}_i,\mathbf{p}^j)}},
\label{eq9}
\end{equation}
where $1 \leq k, j \leq N$ and $k, j \neq y_i$.
    
\begin{figure}[t]
    \centering
        \includegraphics[width=\columnwidth]{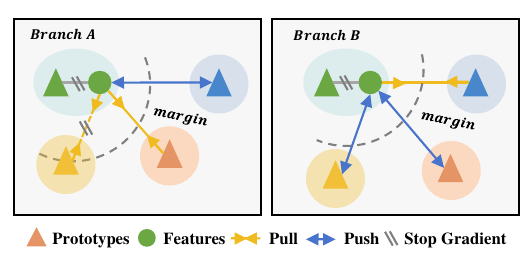}
        \label{lossincon}
        \vspace{-0.5cm}
    \caption{\textbf{An illustration of how inconsistency loss $\mathcal{L}_{incon}$ acts on two branches to adjust the class feature distribution.} The class pairs of each branch are optimized in opposite directions. When the class pairs are pulled within the margin, the optimization of $\mathcal{L}_{incon}$ will halt for one branch.}
    \vspace{-0.3cm}
    \label{fig4}
\end{figure}

The collection of probabilities represents the class feature distribution within each branch, depicted as an $M \times (N-1)$ matrix in \figurename~\ref{fig2}. The computation of class feature distribution involves both sample features and prototypes, ensuring that every sample in the feature space contributes to the adjustment in class feature distribution, not just the representatives (prototypes) of feature clusters. To maximize class feature distribution inconsistency, we propose an \textbf{inconsistency loss} which encourages maximizing the inconsistency of two probabilities as follows: 
\begin{equation}
\begin{aligned}
\mathcal{L}_{incon} = -\frac{1}{M}\sum_{i=1}^M\log&\sum_{k=1}^N({p_{Br_A,i}^k}(1-{p_{Br_B,i}^k}) \\ & + {p_{Br_B,i}^k}(1-{p_{Br_A,i}^k})),
\end{aligned}
\label{eq10}
\end{equation}
where $k \neq y_i$. Here $Br_A$ and $Br_B$ represent two branches, and $p_{Br_A, i}^k$ and $p_{Br_B, i}^k$ represent the respective probability values in the \eqref{eq9}. From a mathematical perspective, the inconsistency loss is minimized when two probability distributions take opposite extreme values. In terms of class feature distribution, minimizing inconsistency loss $\mathcal{L}_{incon}$ narrows the proximity between two classes within one branch while simultaneously increasing the proximity between the corresponding two classes in another branch, as shown in \figurename~\ref{fig4}.

Considering that adjusting the class feature distribution will inevitably pull some features and non-corresponding prototypes close within a branch during training, we introduce the positive distance between features and their corresponding prototypes along with a margin $m_1>0$ in the distance computation of the \eqref{eq8} to mitigate this issue:
\begin{equation}
d({\mathbf{z}_i,\mathbf{p}^j}) = -\mathop{\max}({\mathbf{z}_i}^T \cdot {\mathbf{p}^k} -{\mathbf{z}_i}^T \cdot {\mathbf{p}^j} - m_1, \, 0),
\label{eq11}
\end{equation}
where $1 \leq j \leq N$ and $y_i = k \neq j$. The term ${\mathbf{z}_i}^T \cdot {\mathbf{p}^k}$ is not subject to gradient optimization. The redefined distance metric not only considers the relationship with positive pairs ${\mathbf{z}_i}$ and ${\mathbf{p}^k}$ but also ensures that the inter-class distance does not decrease to within the margin. Specifically, when ${\mathbf{z}_i}^T \cdot {\mathbf{p}^j} > {\mathbf{z}_i}^T \cdot {\mathbf{p}^k} - m_1$, the relative positions of corresponding features and prototypes will not be adjusted in one branch, as shown in \figurename~\ref{fig4}.

\subsection{Individual Inter-class Separability}
\label{sec3.5}
Dual-perspective inconsistency learning encourages differences among perspectives and therefore increase their bias, which may potentially degrade individual performance. While enhancing differences among the dual perspectives, we also focus on maintaining individual performance within each branch. Since individual branches rely on distance metrics for rejection, establishing a decision boundary with effective inter-class separability is crucial. We therefore introduce the triplet loss~\cite{schroff2015facenet} based on prototype learning to optimize the inter-class separability. Triplet loss minimizes the distance between an anchor and a positive, both of which belong to the same class, and minimizes the distance between the anchor and a negative of a different class~\cite{schroff2015facenet}. Neighboring classes naturally form hard negative pairs.

\begin{equation}
\mathcal{L}_{trip} = \frac{1}{M} \sum_{i=1}^M \mathop{\max (\Vert\mathbf{z}_i-\mathbf{p}^k\Vert- \Vert\mathbf{z}_i-\mathbf{p}^j \Vert + m_2, \, 0)},
\label{eq12}
\end{equation}
where $k=y_i$. Class $j$ is the nearest neighbor of class $k$. The loss $\mathcal{L}_{trip}$ pulls the features (anchors) and their corresponding prototypes (positives) closer while pushing them away from the nearest non-corresponding prototypes (negatives), ensuring effective inter-class separability for each individual branch.

The final loss function applied to PredIN is as follows:

\begin{equation}
\mathcal{L}_{Div} = \mathcal{L}_{PL}+\gamma \mathcal{L}_{incon} + \alpha \mathcal{L}_{trip},
\label{eq13}
\end{equation}
where $\gamma$ and $\alpha$ are the weights of $\mathcal{L}_{incon}$ and $\mathcal{L}_{trip}$. Here the inconsistency loss and triplet loss are complementary as the former enhances differences within perspectives by maximizing the class feature distribution inconsistency between branches while the latter maintains the individual performance by optimizing the inter-class separability within each branch. 

\begin{figure}[!t]
    \centering
        \vspace{-0.2cm}
        \includegraphics[width=\columnwidth]{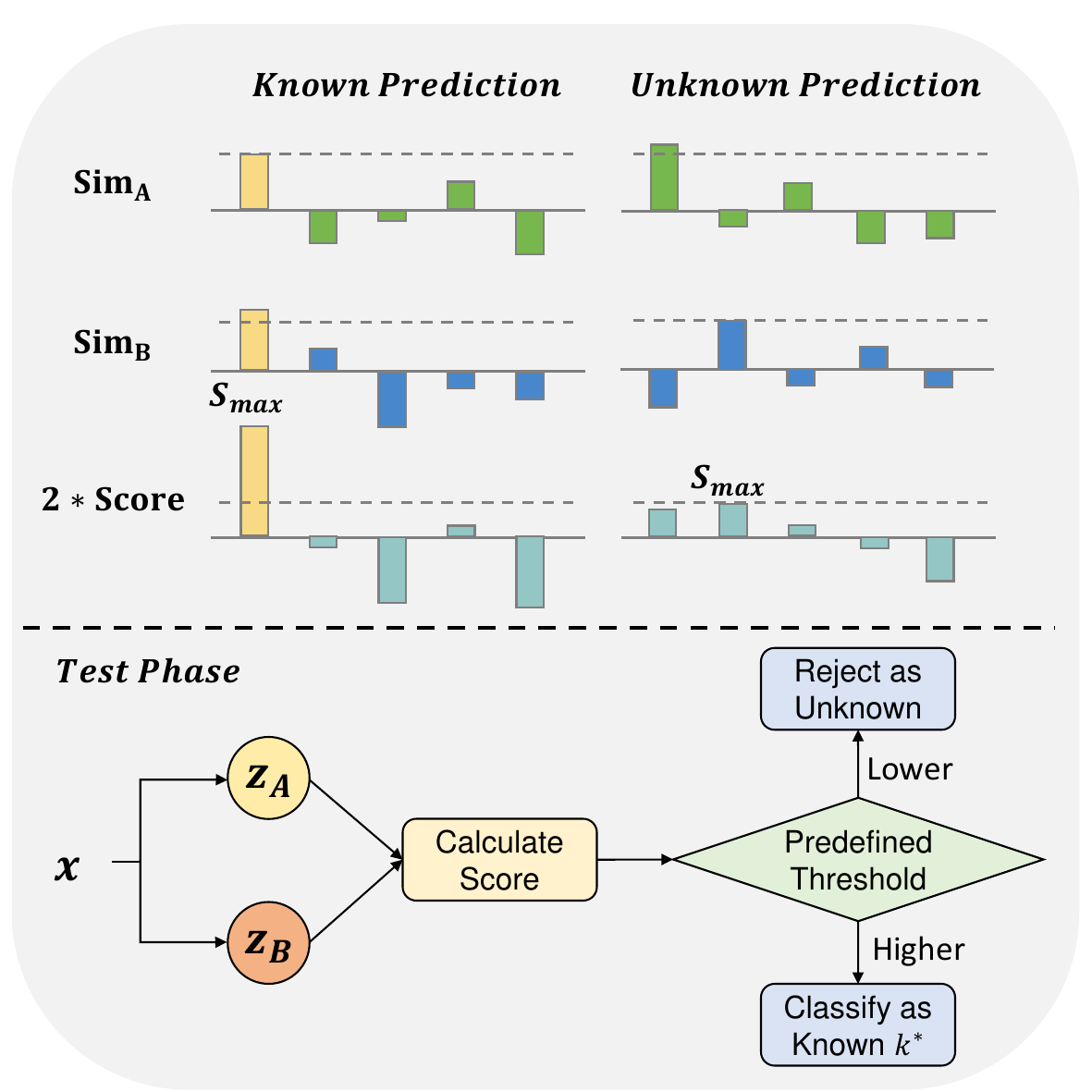}
        \label{rejection rule}
        \vspace{-0.2cm}
    \caption{\textbf{Rejection rules.} Through averaging, unknown samples (right) tend to predict different results and produce lower $S_{max}$, while known samples (left) tend to obtain the same predictions consistent with the label (yellow) and produce higher $S_{max}$.}
    \vspace{-0.5cm}
    \label{fig5}
    
\end{figure}

\subsection{Unknown Rejection}

In PredIN, each branch follows previous PL-based approaches and obtains similarity based on the distance between sample features and prototypes. Specifically, similarity between a given feature $\mathbf{z}_i$ and the prototype $\mathbf{p}^k$ for each branch is defined as follows:
\begin{equation}
\textrm{Sim}(\mathbf{z}_i,\mathbf{p}^k) = {\mathbf{z}_i}^T \cdot \mathbf{p}^{k},
\label{eq14}
\end{equation}
where $1 \leq k \leq N$.

Besides the above distance metric, we also need to address how to leverage prediction inconsistency brought by different perspectives. A commonly used method for integrating predictions from different branches is averaging, which is equally applicable to our task. We combine the outputs of two branches:
\begin{equation}
\textrm{Score}(\mathbf{z}_i,\mathbf{p}^k) =\frac{1}{2}({\textrm{Sim}_A(\mathbf{z}_{i},\mathbf{p}^k}) + {\textrm{Sim}_B(\mathbf{z}_{i},\mathbf{p}^{k})}),
\label{eq15}
\end{equation}
where $1 \leq k \leq N$.

In order to classify sample $\mathbf{x}_i$ as a certain known class $k$ or reject it as unknown classes, we further obtain $S_{max}$:
\begin{equation}
S_{max} =  \textrm{Score}(\mathbf{z}_i,\mathbf{p}^{k^*}),
\label{eq16}
\end{equation}
where
\begin{equation}
k^* = \underset{1 \leq k \leq N}{\arg\max} \ \textrm{Score}(\mathbf{z}_i,\mathbf{p}^{k}).
\label{eq17}
\end{equation}

\begin{table}[!b]
    \caption{Characteristics and setup of four public sEMG datasets. BioDB2, DB2, DB4, and DB7 are abbreviations for BioPat DB2, Ninapro DB2, DB4, and DB7, respectively.} 
    \label{tab1}
    \centering
    \small
    \vspace{-0.1cm}
    \setlength{\tabcolsep}{0.95mm}{
        \begin{tabular}{lcccc}
        \toprule
        Dataset &  BioDB2 & DB2 & DB4 & DB7 \\
        \midrule
        Subjects & 17 & 40 & 10 & 20  \\
        Channels & 8 & 12 & 12 & 12   \\
        Sampling rate & 2000Hz & 2000Hz & 2000Hz & 2000Hz \\
        Trials & 3 & 6 & 6 & 6  \\
        Training Trials & 1,2 & 1,3,4,6 & 1,3,4,6 & 1,3,4,6 \\
        Testing Trials & 3 & 2,5 & 2,5 & 2,5 \\
        Gestures & 27 & 50 & 53 & 41   \\
        Known Gestures & 10 & 15 & 15 & 15 \\
        Unknown Gestures & 17 & 35 & 38 & 26 \\
        \bottomrule
        \end{tabular}}
\end{table}

In summary, we comprehensively consider the distinctions in prediction inconsistency and distance between known and unknown classes, thereby effectively increasing the rejection performance of the unknown. As shown in \figurename~\ref{fig5}, in a single branch, the maximum similarity score of some unknown samples may be higher than that of known samples, making it difficult to reject based on a certain threshold. However, unknown samples will obtain lower $S_{max}$ through averaging due to prediction inconsistency, while known samples will be close to the same-class prototypes among different perspectives resulting in higher $S_{max}$. A pre-determined threshold can be applied to $S_{max}$ to reject the unknown. Samples with $S_{max}$ greater than the threshold value will be regarded as known ones and classified into class $k^*$. Building on the majority of previous OSR studies, a threshold that ensures 95\% of the data is recognized as known is commonly used.

\section{Experiments and results}
\label{sec4}
\subsection{Datasets}

\begin{table*}[!ht]
    \centering
    \caption{Performance comparison with SOTAs in terms of AUC (\%), OSCR (\%) and ACC (\%) on four public datasets. Results are averaged among five randomized splits. Best performances are highlighted in bold.}
    \small
     \vspace{-0.1cm}
    \setlength{\tabcolsep}{0.02cm}{
        \begin{tabular}{lcccccccccccc}
        \toprule
        \multirow{2}[4]{*}{Methods} & \multicolumn{3}{c}{BioPat DB2~}     & \multicolumn{3}{c}{Ninapro DB2} & \multicolumn{3}{c}{Ninapro DB4} & \multicolumn{3}{c}{Ninapro DB7}\\
        \cmidrule(lr){2-4} \cmidrule(lr){5-7} \cmidrule(lr){8-10} \cmidrule(lr){11-13} 
        & ~AUC~ & OSCR & ~ACC~ & ~AUC~ & OSCR & ~ACC~ & ~AUC~ & OSCR & ~ACC~ & ~AUC~ & OSCR & ~ACC~ \\
        \midrule
        Softmax  & 65.5 & 60.5 & 86.0 & 72.2 & 65.6 & 82.6 & 73.1 & 61.7 & 75.3 & 73.5 & 66.8 & 83.0 \\
        PROSER (CVPR'21) & 67.9 & 63.5 & 87.8 & 72.4 & 63.9 & 80.0 & 74.6 & 63.1 & 76.1 & 73.5 & 66.7 & 82.9 \\
        CAC (WACV'21) & 68.6 & 63.9 & 87.7 & 72.9 & 65.8 & 82.5 & 75.5 & 63.9 & 76.4 & 72.6 & 66.3 & 83.7 \\
        CPN-MGR (IEEE Sens. J'22) & 69.7 & 63.3 & 85.9 & 76.4 & 68.5 & 82.6 & 76.6 & 63.2 & 75.0 & 76.4 & 68.4 & 83.1 \\
        ARPL (TPAMI'22)   & 68.2 & 61.1 & 85.7 & 76.3 & 67.9 & 82.7 & 79.6 & 64.5 & 74.8 &79.1 & 70.1 & 82.9\\
        SLCPL (CVIU'23) & 70.1 & 63.7 & 85.8 & 76.5 & 68.4 & 82.5 & 76.4 & 63.3 & 75.5 & 76.6 & 68.4 & 82.8 \\
        DIAS (ECCV'22)  & 70.2 & 65.8 & 88.1 & 77.8 & 69.7 & 83.5 & 80.2 & 65.7 & 75.9 & 79.8 & 71.0 & 83.5 \\
        \midrule
        \textbf{PredIN (Crossformer)} & \textbf{74.9} & \textbf{69.8} & \textbf{89.2} &  \textbf{80.4} & \textbf{72.5} & \textbf{85.2} & \textbf{80.8} & \textbf{68.2} & \textbf{78.1}  & \textbf{81.0} & \textbf{73.1} & \textbf{85.2} \\
        \midrule
        OpenMax (CVPR'16) & 68.6 & 64.7 & 87.8 & 69.2 & 58.4 & 74.7 & 74.5 & 61.2 & 73.0 & 71.1 & 62.2 & 78.2\\
        PROSER (CVPR'21) & 75.0 & 72.6 & \textbf{93.5} & 69.6 & 56.1 & 71.4 & 76.2 & 59.3 & 69.9  & 73.4 & 63.6 & 78.7 \\
        CAC (WACV'21) & 72.2 & 69.9 & 93.1 & 70.4 & 58.8 & 74.7 & 76.8 & 63.1 & 73.8 & 73.2 & 64.0 & 79.5 \\
        CSSR (TPAMI'22) & 70.0 & 65.1 & 92.4 & 70.3 & 56.9 & 72.7 & 77.7 & 57.5 & 73.5 & 73.3 & 56.7 & 78.0 \\
        CPN-MGR (IEEE Sens. J'22) & 73.0 & 69.3 & 92.0 & 71.9 & 55.4 & 69.5 & 79.3 & 58.0 & 67.7 & 78.2 & 63.6 & 75.3\\
        ARPL (TPAMI'22) & 73.4 & 70.0 & 91.9 & 72.2 & 59.1 & 74.7 & 79.8 & 63.0 & 73.1 & 78.6 & 64.6 & 78.2 \\
        MGPL (Inf. Sci'23) & 70.6 & 67.3 & 91.5 & 62.3 & 44.7 & 65.4 & 76.7 & 58.3 & 70.1 & 64.9 & 51.7 & 68.9\\
        SLCPL (CVIU'23) & 72.1 & 69.5 & 93.2 & 71.5 & 57.3 & 75.8 & 77.9 & 61.8 & 72.3 & 77.1 & 64.1 & 76.8\\
        MEDAF (AAAI'24) & 74.7 & 72.2 & 93.0 & 72.9 & 63.2 & \textbf{77.7} & 80.2 & 64.9 & 74.4 & 79.9 & 70.5 & \textbf{82.4}\\
        \midrule
        \textbf{PredIN (Hybrid model)} & \textbf{75.7} & \textbf{72.7} & 93.4 & \textbf{76.7} &\textbf{64.3} & 77.6 & \textbf{82.2} &\textbf{67.6} & \textbf{76.9} & \textbf{82.1} & \textbf{71.6} & 82.2\\
        \bottomrule
        \end{tabular}}
        \label{tab2}
        \vspace{-0.2cm}
\end{table*}

We apply four public sEMG benchmark datasets~\cite{OrtizCataln2013BioPatRecAM, DB2, DB4, DB7} to validate the proposed approach, as shown in Table~\ref{tab1}. During preprocessing, raw sEMG signals are segmented via a sliding window of length 200 ms with steps of 50 ms, and then standardized channel-wise. As recommended by BioPatRec~\cite{OrtizCataln2013BioPatRecAM}, we remove transient periods of the contraction using a contraction time percentage of $0.7$ for the BioPat DB2 dataset. According to the setting of closed-set gesture recognition based on sEMG~\cite{karnam2022emghandnet, wang2024transformer}, the training and testing set are divided based on trials as mentioned in Table~\ref{tab1}. Following the protocol of open-set image recognition~\cite{Chen2022ARPL}, we randomly select $10$ known classes from BioPat DB2 and $15$ known classes from Ninapro DB2, Ninapro DB4 and Ninapro DB7, while treating the remaining classes as unknown.

\subsection{Evaluation Metrics}

We use three common metrics to measure the performance of OSR derived from~\cite{Chen2022ARPL, park2024understanding}: (1) the area under the receiver operating characteristic (AUC); (2) closed-set classification accuracy (ACC); (3) open-set classification rate (OSCR). They are all threshold-independent metrics. Further details are provided as follows:

\begin{itemize}
    \item [$\bullet$] \textbf{AUC} measures the model's ability to distinguish between known and unknown classes based on the relationship between true positive rate (TPR) and false positive rate (FPR).
    \item [$\bullet$] \textbf{ACC} assesses known classes classification performance.
    \item [$\bullet$] \textbf{OSCR} comprehensively evaluates empirical classification risk and open space risk based on closed-set classification accuracy (ACC) and false positive rate (FPR).
\end{itemize}

\subsection{Experimental Settings}
To comprehensively extract sEMG features, we employ two types of encoders, the Crossformer~\cite{zhang2022crossformer} and a CNN-LSTM hybrid network, as the backbone. The Crossformer is a popular Transformer-based model for time series forecasting while also showing superiority in sEMG classification tasks since it effectively captures the cross-time and cross-channel dependency and extracts multi-scale time information. We set the segment length of Crossformer as $32$, window size as $2$, and layers as $5$. In addition, we design a hybrid network based on CNN and LSTM, which combines a ResNet variant, an LSTM and an SKAttention module.

During experiments, we use the SGD optimizer with an initial learning rate of $0.01$. The learning rate decreases by a factor of $0.1$ at $60$ and $80$ epochs. The batch size is set to $256$ and the training epoch is set to $100$.
The hyperparameters $\beta$, $\gamma$ and $\alpha$ in \eqref{eq7} and \eqref{eq13} are all empirically set to $1.0$, while the feature dimension of embedding space is set to $128$. Two margins in \eqref{eq11} and \eqref{eq12} are set to $0.5$ and $1.0$ respectively.
Prototypes are randomly initialized by the standard normal distribution. All experimental results are averaged among five randomized splits of datasets by classes, which means each split uses different known classes to train. The PredIN is implemented by using Pytorch 2.3.0 and executed on an NVIDIA GeForce RTX 4090 GPU.

\subsection{Comparison with the State-of-the-arts}

We compare our method against other state-of-the-art open-set image and gesture recognition approaches. Softmax and OpenMax are methods based on the prediction probability. OpenMax~\cite{Bendale2016Openmax} uses extreme value theory (EVT) to calibrate the prediction probability. PROSER~\cite{Zhou2021Palceholder} augmented the closed-sed classifier with an extra classifier placeholder and mimics new data with manifold mixup. CAC~\cite{millerclass} proposes a new distanced-based loss function to encourage clustering among known classes. CSSR~\cite{huang2022class} replaces prototype points with manifolds represented by class-specific auto-encoders(AE). ARPL~\cite{Chen2022ARPL} and SLCPL~\cite{xia2023slcpl} are methods based on prototype learning and design various distance loss functions to reduce open space risk. DIAS~\cite{moon2022difficulty} considers different difficulty levels and introduces an image generator and a feature generator to produce hard fake instances. MGPL~\cite{liu2023MGPL} is also the method based on PL but applies the VAE framework to optimize generative constraints. MEDAF~\cite{wang2024exploring} learns diverse representation with an attention diversity regularization. CPN-MPR~\cite{Wu2022UnknownMR} focuses on open-set sEMG-based gesture recognition and introduces the PL to reject the unknown. To ensure the fairness of the comparison, all methods employ the same backbone.

The results in Table~\ref{tab2} highlight the performance of our proposed approach for the open-set sEMG-based gesture recognition task, even though across different backbone architectures. Specifically, considering the rejection performance, our method achieves the best AUC scores of {\bf75.7\%}, {\bf76.7\%}, {\bf82.2\%} and {\bf82.1\%} across four datasets. Moreover, dual-perspective inconsistency learning brings benefits to closed-set classification tasks. Our method achieves significant improvements in closed-set accuracy compared to these SOTA methods. Furthermore, when considering both empirical classification risk and open space risk, our approach also surpasses the above SOTA methods, consistently achieving the highest OSCR scores on four datasets. These results confirm that prediction inconsistency reveals the distinctions between known and unknown classes effectively. In conclusion, our approach shows the superiority in both closed-set classification and unknown rejection.

\begin{table*}[!ht]
    \centering
    \caption{Ablations of each module in terms of AUC (\%), OSCR (\%) and ACC (\%) on four public datasets. Best performances are highlighted in bold.}
    \small
    \vspace{-0.1cm}
    \setlength{\tabcolsep}{0.05cm}{
        \begin{tabular}{lccccccccccccc}
        \toprule
        \multirow{2}[4]{*}{Methods} & \multicolumn{3}{c}{BioPat DB2}     & \multicolumn{3}{c}{Ninapro DB2} & \multicolumn{3}{c}{Ninapro DB4} & \multicolumn{3}{c}{Ninapro DB7}\\
        \cmidrule(lr){2-4} \cmidrule(lr){5-7} \cmidrule(lr){8-10} \cmidrule(lr){11-13} 
        & ~AUC~ & OSCR & ~ACC~ &~AUC~ & OSCR & ~ACC~ & ~AUC~ & OSCR & ~ACC~& ~AUC~ & OSCR & ~ACC~\\
        \midrule
        PL baseline & 71.5 & 66.5 & 89.1 & 73.3 & 59.1 & 73.2 & 79.1 & 61.7 & 71.9 & 77.7 & 65.4 & 77.7\\
        Dual-perspective   & 73.2 & 68.9 & 90.4 & 76.0 & 64.0 & 77.5 & 80.9 & 65.8 & 75.5 & 80.2 & 70.2 & 81.8\\
        Dual-perspective (w/ $\mathcal{L}_{trip}$)  & 73.8 & 70.7 & 92.3 & 76.4 & 64.2 & 77.4 & 81.8 & 67.4 & 76.9 & 80.8 & 71.0 & 82.2 \\
        PredIN (w/o $\mathcal{L}_{trip}$)& 72.4 & 68.3 & 90.3 & 75.6 & 61.9 & 75.6 & 81.6 & 66.4 & 75.6 & 81.1 & 70.6 & 81.8\\
        \midrule
        PredIN & \textbf{75.7} & \textbf{72.7} & \textbf{93.4} & \textbf{76.7} & \textbf{64.3} & \textbf{77.6} & \textbf{82.2} & \textbf{67.6} & \textbf{76.9} & \textbf{82.1} & \textbf{71.6} & \textbf{82.2}\\
        \bottomrule
        \end{tabular}}
        \label{tab4}
\end{table*}

\begin{figure*}[!t]
    \centering
    \begin{minipage}[b]{\columnwidth}
        \centering
        \includegraphics[width=0.8\columnwidth]{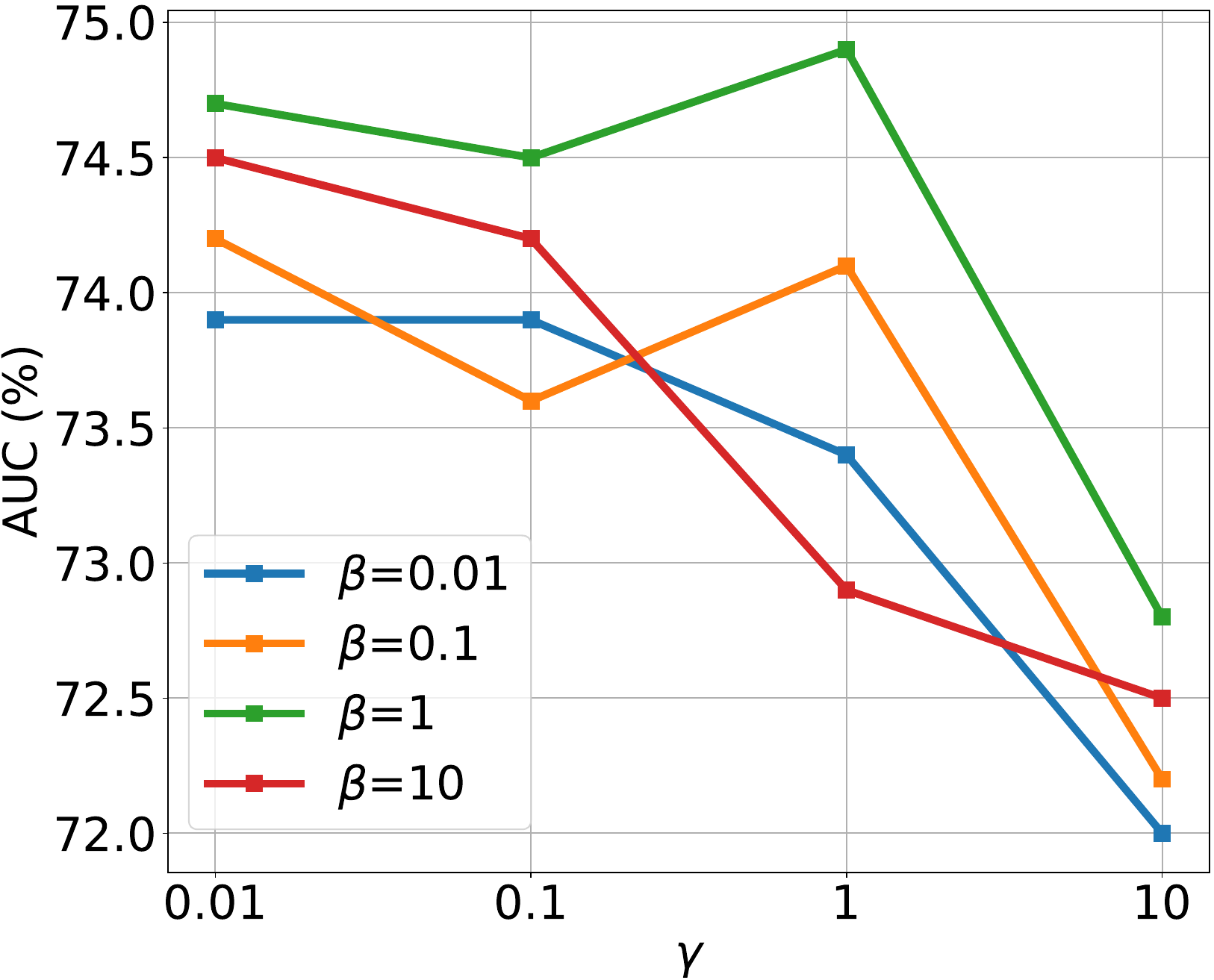}
    \end{minipage}
    \begin{minipage}[b]{\columnwidth}
        \centering
         \hspace{-1.2cm}
        \includegraphics[width=0.8\columnwidth]{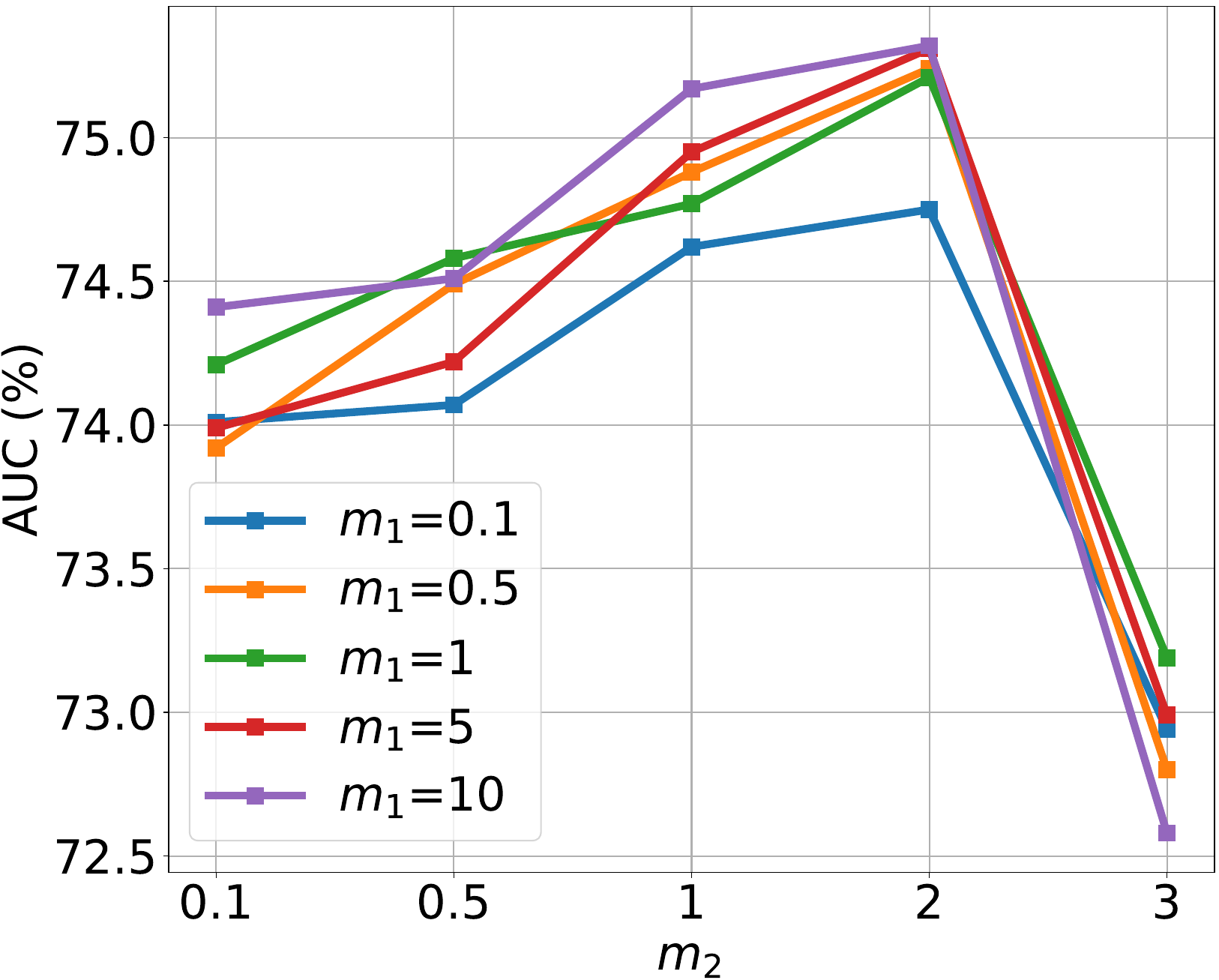}
    \end{minipage}
    \vspace{-0.2cm}
    \caption{\textbf{Evaluation on hyperparamers impact.} (a) The AUC performance under different weights $\beta$ and $\gamma$. (b) The AUC performance under different margins $m_1$ and $m_2$.}
    \vspace{-0.4cm}
    \label{fig6}
\end{figure*}

\begin{figure*}[ht]
    \centering
    \begin{minipage}[b]{0.9\textwidth}
        \centering
        \includegraphics[width=0.9\textwidth]{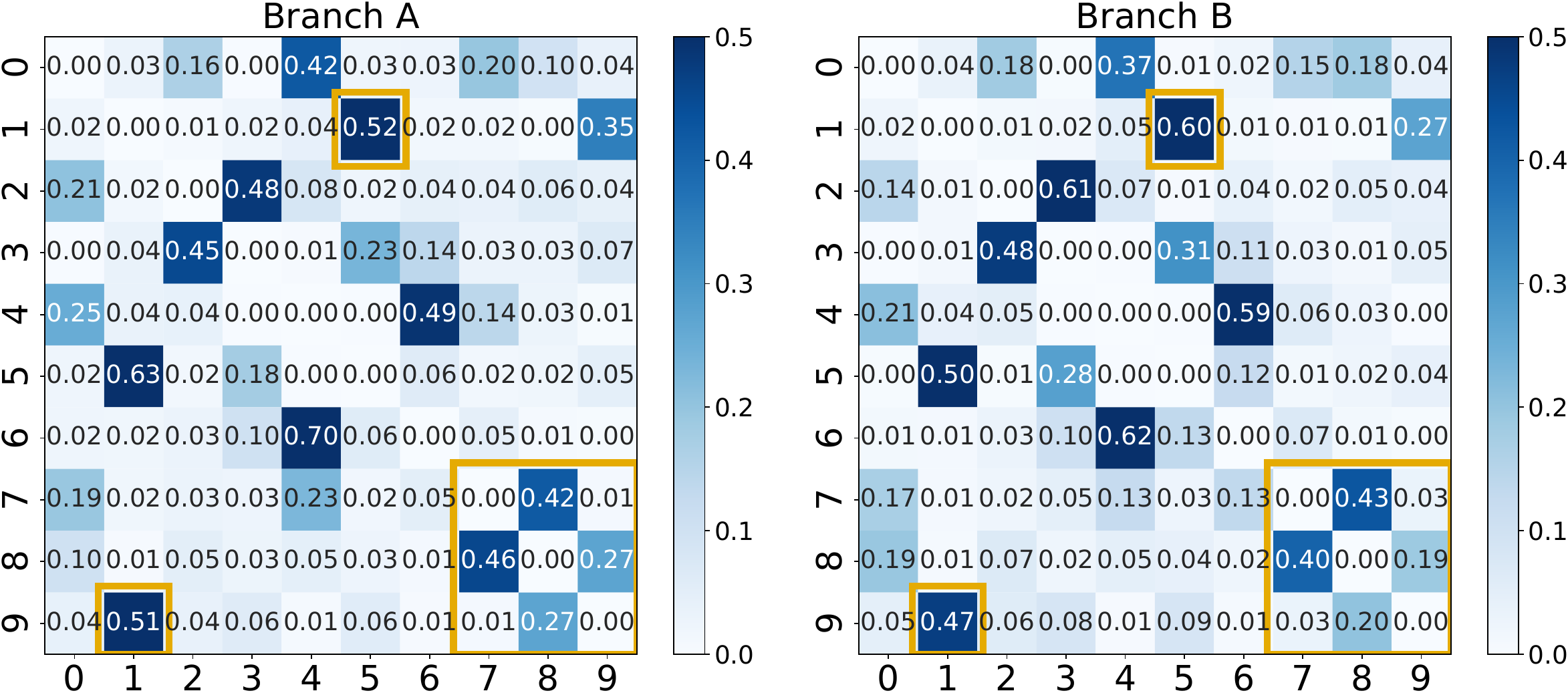}
        \caption*{(a) Dual-perspective}
        \label{fig7:a}
    \end{minipage}
    \vspace{0.7cm} 
    \begin{minipage}[b]{0.9\textwidth}
        \centering
        \includegraphics[width=0.9\textwidth]{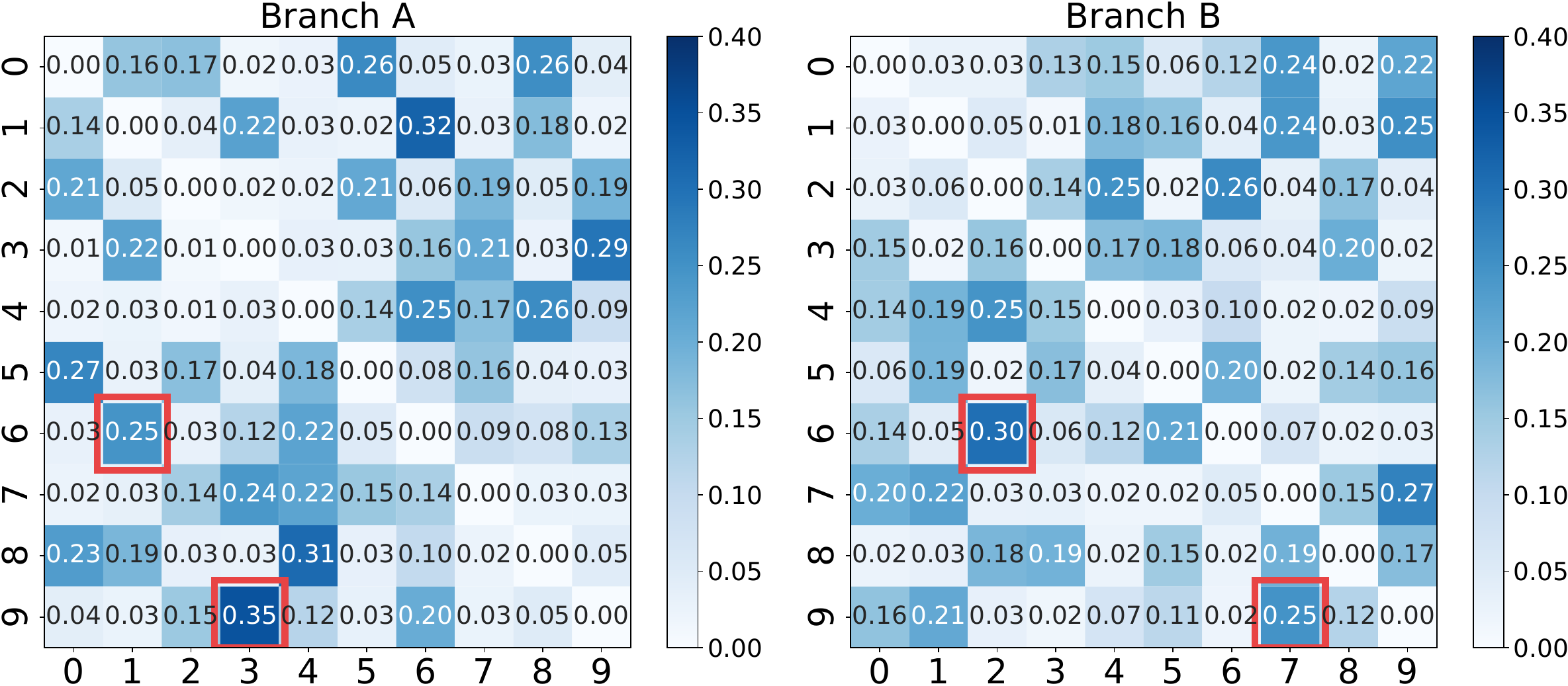}
        \caption*{(b) PredIN}
        \label{fig7:b}
    \end{minipage}
    \vspace{-0.9cm}
    \caption{\textbf{The visualization of class feature distribution of dual perspectives.} Both horizontal and vertical coordinates represent class labels. Each value represents the degree of proximity between two classes within a single branch. (a) represents the two branches of the Dual-perspective model. There are similar local structures between the two branches (yellow boxes). (b) represents the two branches of the PredIN model. The global layout and local neighboring pairs are different (red boxes).} 
    \vspace{-0.3cm}
    \label{fig7}
\end{figure*}

\subsection{Ablation Study}

{\bf Module Ablation.} As presented in Table~\ref{tab4}, each component within our method has been systematically integrated into the PL baseline to verify its necessity. 
We first introduce the Dual-perspective framework, which combines two identical networks with different random initializations. As shown in the second raw, the introduction of dual perspectives yields notable improvements in AUC and OSCR scores compared to the PL baseline, demonstrating the effectiveness of using prediction inconsistency for unknown rejection. Secondly, we further verify the effectiveness of our proposed two losses. In PredIN, the roles of inconsistency loss and triplet loss are complementary: the loss $\mathcal{L}_{trip}$ acts to optimize the inter-class separability within each branch while the inconsistency loss $\mathcal{L}_{incon}$ maximizes the class feature distribution inconsistency among different perspectives. To further explain, we apply the loss $\mathcal{L}_{trip}$ to the Dual-perspective model alone and then combine these two losses into the model training. The standalone application of $\mathcal{L}_{trip}$ enhances the individual rejection capability, thereby consistently improving overall performance compared to the Dual-perspective model. When combining these two losses, PredIN magnifies the prediction inconsistency by enhancing the differences between dual perspectives while maintaining the individual performance, which provides further improvements and clearly demonstrates the effectiveness of maximizing the class feature distribution inconsistency. In addition, we remove the loss $\mathcal{L}_{trip}$ from the PredIN to verify the importance of maintaining individual performance from another aspect. A clear decrease occurs on BioPat DB2 and Ninapro DB2 after the removal.
Finally, incorporating all the above components improves AUC by an average of {\bf +3.8\%} compared to the baseline PL model, which demonstrates that each component contributes to the overall improvement on unknown rejection.

{\bf Hyperparameters Ablation.} We evaluate the effect of two sets of hyperparameters: the trade-off weights $\beta$ and $\gamma$ in \eqref{eq7} and \eqref{eq13}, and two margins $m_1$ and $m_2$ in \eqref{eq11} and \eqref{eq12}. These experiments are conducted using the bioDB2 dataset. \figurename~\ref{fig6} shows the AUC for different $\beta$ and $\gamma$ values. The weight $\beta$ influences the compactness of feature space in prototype learning. According to the results, larger values yield better results, but excessively large values may cause optimization issues and lead to a performance decrease. The weight $\gamma$ represents the degree of adjustment to the class feature distribution. Larger values similarly affect the model's classification. Setting $\beta$ and $\gamma$ to $1.0$ is optimal. Two margins influence the degree of class feature distribution inconsistency and inter-class separability. Greater values of $m_1$ and $m_2$ lead to reduced inconsistency but increased enhancement for inter-class separability. A balanced result is that the optimal values for $m_1$ and $m_2$ are set to $0.5$ and $1.0$ in our experiments, respectively.

\section{Further Analysis and Discussion}
\label{sec5}

\subsection{Dual-perspective Prediction Inconsistency Analysis}
The improvement in performance compared to the Dual-perspective model confirms the increased prediction inconsistency introduced by PredIN. To further evaluate this improvement, we visualize the class feature distribution and measure the prediction inconsistency with an inconsistency metric $Incon$.

We first verify that our approach enlarges the differences between perspectives by maximizing the class feature distribution inconsistency. Specifically, we visualize the class feature distribution using a proximity matrix. As the class feature distribution can be approximately characterized by learned prototypes, we compute the distances between class prototypes and convert them into probabilities in order to represent class proximity. \figurename~\ref{fig7} demonstrates that PredIN achieves our desired class feature distribution inconsistency between two branches, especially compared to the Dual-perspective model. Locally, each class has different neighboring classes, while globally, the relative positions of classes vary, which causes the feature spaces of the two perspectives different.

\begin{figure}[t]
    \centering
        \includegraphics[width=0.95\columnwidth]{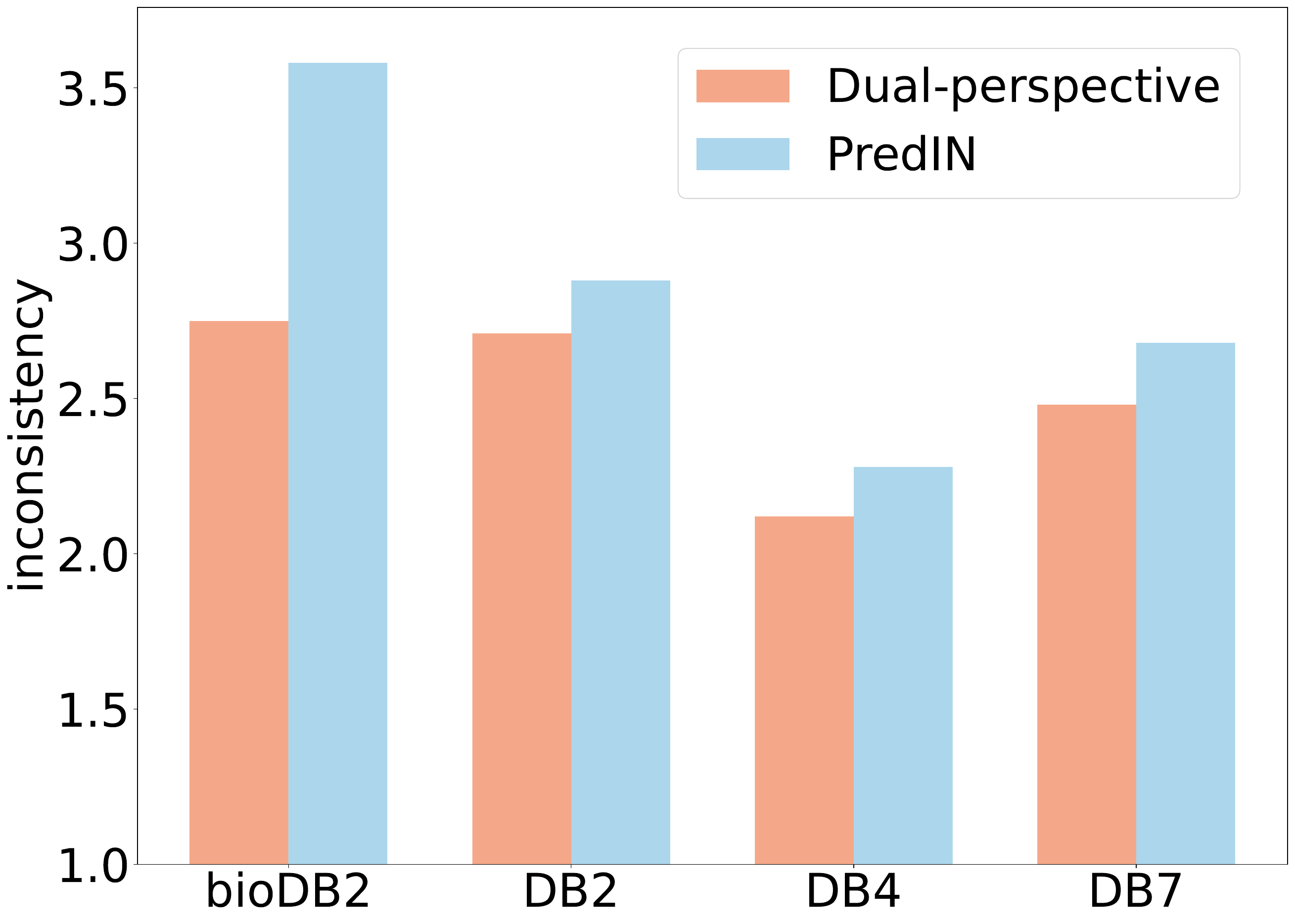}
    \vspace{-0.2cm}
    \caption{\textbf{Inconsistency comparison on four sEMG datasets.}}
    \vspace{-0.4cm}
    \label{fig8}
\end{figure}

To further validate that maximizing class feature distribution inconsistency enhances prediction inconsistency for unknown samples between two perspectives, we compare the PredIN with the Dual-perspective model using an inconsistency metric. We focus on the fraction of prediction changes both in known and unknown samples to measure the relative prediction inconsistency between perspectives on unknown samples:
\begin{equation}
Incon = \frac{Fraction \ of \ unknown \ prediction \ changes}{Fraction \ of \ known \ prediction \ changes}.
\label{eq18}
\end{equation}

The results in \figurename~\ref{fig8} demonstrate the consistent prediction inconsistency improvements achieved by PredIN across four sEMG datasets.

\subsection{Perspective Number}
\begin{figure}[t]
    \centering
    \begin{minipage}[b]{\columnwidth}
        \centering
        \includegraphics[width=0.75\columnwidth]{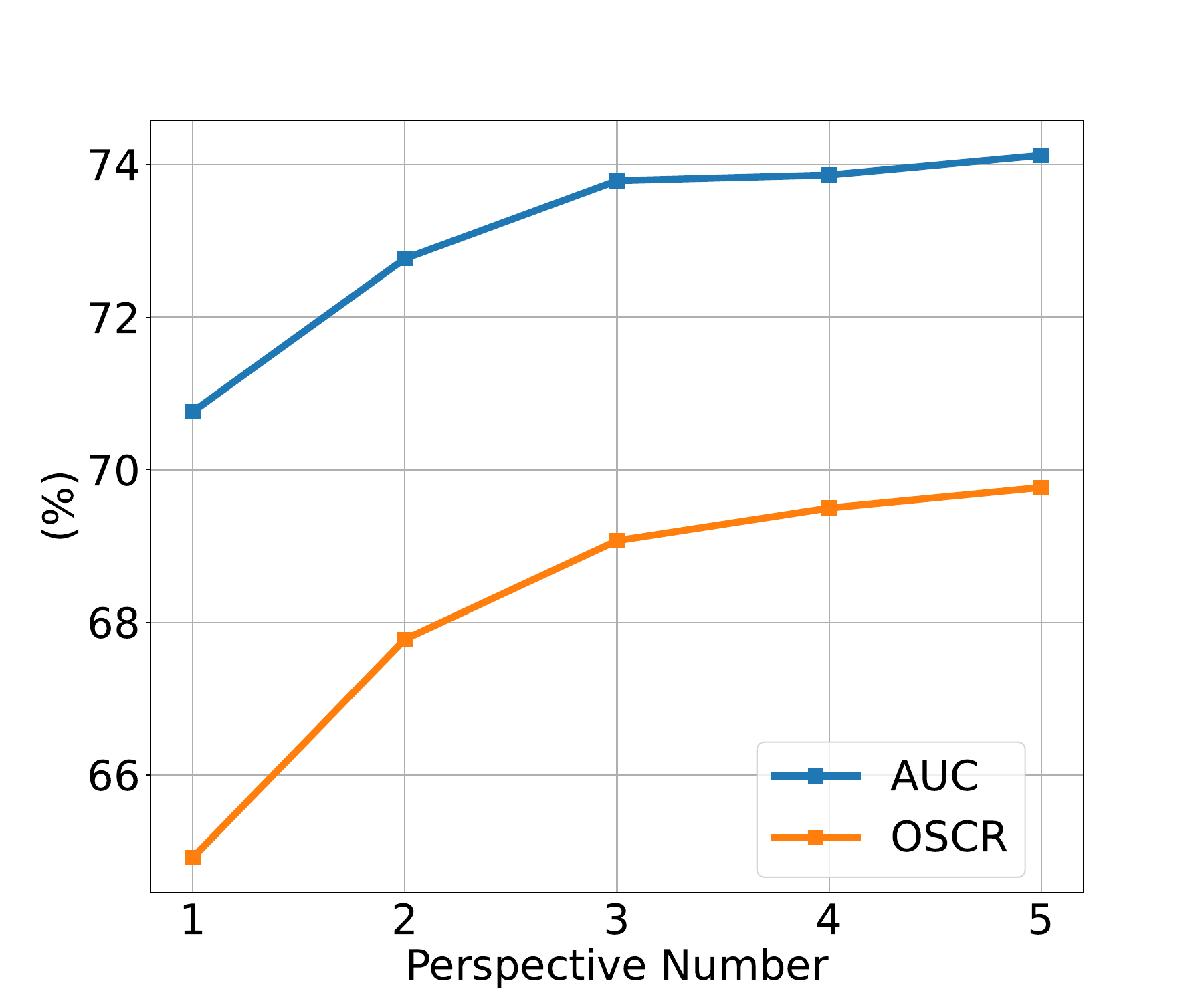}
        \vspace{0.5cm}
    \end{minipage}
    \begin{minipage}[b]{\columnwidth}
        \centering
        \includegraphics[width=0.75\columnwidth]{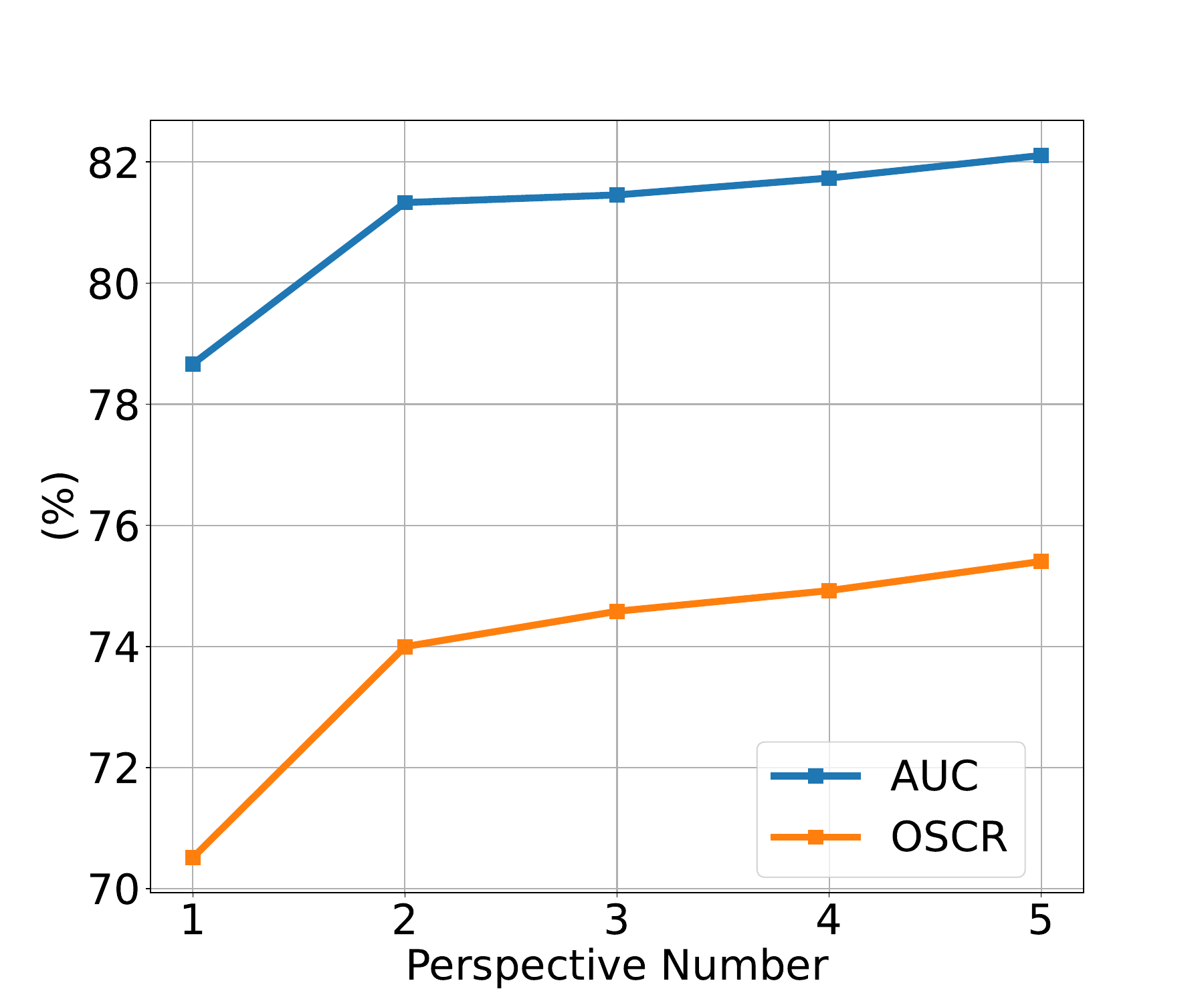}
    \end{minipage}
    \vspace{-0.2cm}
    \caption{\textbf{Evaluation on the impact of number of perspectives.} (a) BioPat DB2. (b) Ninapro DB7.}
    \label{fig10}
    \vspace{-0.5cm}
\end{figure}

In our experiments, we use two perspectives because the inconsistency loss has a symmetrical form, which makes it unsuitable for optimizing more branches in parallel. To evaluate whether more perspectives will perform better, we train multiple models sequentially. The training of the first model is only based on the PL loss $\mathcal{L}_{PL}$, minimizing its empirical classification risk. Subsequent models apply our proposed approach and are optimized based on the loss $\mathcal{L}_{Div}$. Each model obtains a different class feature distribution from the former. \figurename~\ref{fig10} shows the results for a larger perspective number of $5$ on BioPat DB2 and Ninapro DB7. The more perspectives gain better performance, which aligns with the prediction inconsistency among different perspectives. However, combining more perspectives will lead to performance saturation and increased computational burden.

\subsection{Limitations and Future Work}

Although our proposed PredIN addresses the challenge of open-set gesture recognition and demonstrates promising performance, it has several limitations. One limitation is the computation cost associated with the dual-perspective framework. Dual-perspective inconsistency learning brings prediction inconsistency but also introduces certain computational costs. In future work, we can mitigate this by sharing a shallow encoder among branches and can combine multiple perspectives. Additionally, enhancing the differences among multiple perspectives and improving the performance of individual branches are orthogonal objectives. By developing an individual branch with optimal performance and combining it with the differences of the multi-perspective model, we can further enhance the model's OSR performance. The potential of multi-perspective models in OSR tasks remains to be fully explored. Moreover, our approach is limited to the rejection of unknown gestures, without further considering the system's ability to learn and classify novel gestures dynamically. One potential future work is to incorporate open-world recognition frameworks, such as leveraging incremental learning techniques.

\section{CONCLUSION}

\label{sec6}

Generalizing gesture recognition from closed-set to open-set is important for real-world HMI. To tackle open-set gesture recognition based on sEMG, we propose a novel dual-perspective inconsistency learning approach, PredIN, based on our observed pronounced prediction inconsistency within different perspectives for the unknown. Specifically, we propose two complementary strategies to improve OSR performance by enhancing perspective differences while maintaining individual performance. Extensive experiments conducted on multiple public datasets consistently demonstrate that our approach outperforms previous state-of-the-art open-set classifiers. Compared to existing methods for closed-set gesture recognition systems, our approach can maintain high classification accuracy for predefined gestures, while effectively rejecting gestures of disinterest. We hope this work will boost the applications of gesture recognition technologies in real-world scenarios. 
Moreover, this research offers new insights into open-set recognition and has great potential to be extended to various fields. Moving forward, we will also explore extending our technology to adapt to diverse pattern recognition tasks.


\bibliographystyle{IEEEtran}
\bibliography{IEEEabrv,refs}

\begin{thebibliography}{10}
\providecommand{\url}[1]{#1}
\csname url@samestyle\endcsname
\providecommand{\newblock}{\relax}
\providecommand{\bibinfo}[2]{#2}
\providecommand{\BIBentrySTDinterwordspacing}{\spaceskip=0pt\relax}
\providecommand{\BIBentryALTinterwordstretchfactor}{4}
\providecommand{\BIBentryALTinterwordspacing}{\spaceskip=\fontdimen2\font plus
\BIBentryALTinterwordstretchfactor\fontdimen3\font minus \fontdimen4\font\relax}
\providecommand{\BIBforeignlanguage}[2]{{%
\expandafter\ifx\csname l@#1\endcsname\relax
\typeout{** WARNING: IEEEtran.bst: No hyphenation pattern has been}%
\typeout{** loaded for the language `#1'. Using the pattern for}%
\typeout{** the default language instead.}%
\else
\language=\csname l@#1\endcsname
\fi
#2}}
\providecommand{\BIBdecl}{\relax}
\BIBdecl

\bibitem{dere2023event}
M.~D. Dere, J.-H. Jo, and B.~Lee, ``Event-driven edge deep learning decoder for real-time gesture classification and neuro-inspired rehabilitation device control,'' \emph{IEEE Transactions on Instrumentation and Measurement}, 2023.

\bibitem{tang2024exosuit}
Z.~Tang, Z.~Zhu, S.~Lv, X.~Hong, Y.~Peng, and N.~Chen, ``An exosuit system with bidirectional hand support for bilateral assistance based on dynamic gesture recognition,'' \emph{IEEE Transactions on Neural Systems and Rehabilitation Engineering}, 2024.

\bibitem{kang2022reduce}
P.~Kang, J.~Li, S.~Jiang, and P.~B. Shull, ``Reduce system redundancy and optimize sensor disposition for emg--imu multimodal fusion human--machine interfaces with xai,'' \emph{IEEE Transactions on Instrumentation and Measurement}, vol.~72, pp. 1--9, 2022.

\bibitem{islam2022application}
M.~J. Islam, S.~Ahmad, F.~Haque, M.~B.~I. Reaz, M.~A.~S. Bhuiyan, and M.~R. Islam, ``Application of min-max normalization on subject-invariant emg pattern recognition,'' \emph{IEEE Transactions on Instrumentation and Measurement}, vol.~71, pp. 1--12, 2022.

\bibitem{islam2024surface}
M.~R. Islam, D.~Massicotte, P.~Massicotte, and W.-P. Zhu, ``Surface emg-based inter-session/inter-subject gesture recognition by leveraging lightweight all-convnet and transfer learning,'' \emph{IEEE Transactions on Instrumentation and Measurement}, 2024.

\bibitem{xiong2024patchemg}
B.~Xiong, W.~Chen, H.~Li, Y.~Niu, N.~Zeng, Z.~Gan, and Y.~Xu, ``Patchemg: Few-shot emg signal generation with diffusion models for data augmentation to improve classification performance,'' \emph{IEEE Transactions on Instrumentation and Measurement}, 2024.

\bibitem{Scheirer2013Open}
W.~J. Scheirer, A.~de~Rezende~Rocha, A.~Sapkota, and T.~E. Boult, ``Toward open set recognition,'' \emph{IEEE Transactions on Pattern Analysis and Machine Intelligence}, vol.~35, no.~7, pp. 1757--1772, 2013.

\bibitem{wu2021metric}
L.~Wu, X.~Zhang, X.~Zhang, X.~Chen, and X.~Chen, ``Metric learning for novel motion rejection in high-density myoelectric pattern recognition,'' \emph{Knowledge-Based Systems}, vol. 227, p. 107165, 2021.

\bibitem{Wu2022UnknownMR}
L.~Wu, A.~Liu, X.~Zhang, X.~Chen, and X.~Chen, ``Unknown motion rejection in myoelectric pattern recognition using convolutional prototype network,'' \emph{IEEE Sensors Journal}, vol.~22, pp. 4305--4314, 2022.

\bibitem{furui2021emg}
A.~Furui, T.~Igaue, and T.~Tsuji, ``Emg pattern recognition via bayesian inference with scale mixture-based stochastic generative models,'' \emph{Expert Systems with Applications}, vol. 185, p. 115644, 2021.

\bibitem{huang2022class}
H.~Huang, Y.~Wang, Q.~Hu, and M.~Cheng, ``Class-specific semantic reconstruction for open set recognition,'' \emph{IEEE Transactions on Pattern Analysis and Machine Intelligence}, vol.~45, no.~4, pp. 4214--4228, 2022.

\bibitem{park2024understanding}
J.~Park, H.~Park, E.~Jeong, and A.~B.~J. Teoh, ``Understanding open-set recognition by jacobian norm and inter-class separation,'' \emph{Pattern Recognition}, vol. 145, p. 109942, 2024.

\bibitem{Yang2020ConvolutionalPN}
H.~Yang, X.~Zhang, F.~Yin, Q.~Yang, and C.~Liu, ``Convolutional prototype network for open set recognition,'' \emph{IEEE Transactions on Pattern Analysis and Machine Intelligence}, vol.~44, no.~5, pp. 2358--2370, 2022.

\bibitem{Chen2022ARPL}
G.~Chen, P.~Peng, X.~Wang, and Y.~Tian, ``Adversarial reciprocal points learning for open set recognition,'' \emph{IEEE Transactions on Pattern Analysis and Machine Intelligence}, vol.~44, no.~11, pp. 8065--8081, 2022.

\bibitem{rame2021dice}
A.~Ram{\'e} and M.~Cord, ``Dice: Diversity in deep ensembles via conditional redundancy adversarial estimation,'' in \emph{ICLR}, 2021.

\bibitem{Xiong2021Review}
D.~Xiong, D.~Zhang, X.~Zhao, and Y.~Zhao, ``Deep learning for emg-based human-machine interaction: {A} review,'' \emph{IEEE/CAA Journal of Automatica Sinica}, vol.~8, no.~3, pp. 512--533, 2021.

\bibitem{Park2016MovementID}
K.~Park and S.~Lee, ``Movement intention decoding based on deep learning for multiuser myoelectric interfaces,'' in \emph{BCI}, 2016, pp. 1--2.

\bibitem{DB2}
M.~Atzori, A.~Gijsberts, C.~Castellini, B.~Caputo, A.-G.~M. Hager, S.~Elsig, G.~Giatsidis, F.~Bassetto, and H.~M{\"u}ller, ``Electromyography data for non-invasive naturally-controlled robotic hand prostheses,'' \emph{Scientific Data}, vol.~1, 2014.

\bibitem{karnam2022emghandnet}
N.~K. Karnam, S.~R. Dubey, A.~Turlapaty, and B.~Gokaraju, ``Emghandnet: A hybrid cnn and bi-lstm architecture for hand activity classification using surface emg signals,'' \emph{Biocybernetics and Biomedical Engineering}, 2022.

\bibitem{liu2024transformer}
Y.~Liu, X.~Li, L.~Yang, and H.~Yu, ``A transformer-based gesture prediction model via semg sensor for human-robot interaction,'' \emph{IEEE Transactions on Instrumentation and Measurement}, 2024.

\bibitem{sun2023OSRsurvey}
J.~Sun and Q.~Dong, ``A survey on open-set image recognition,'' \emph{arXiv preprint arXiv:2312.15571}, 2023.

\bibitem{Bendale2016Openmax}
A.~Bendale and T.~E. Boult, ``Towards open set deep networks,'' in \emph{CVPR}, 2016, pp. 1563--1572.

\bibitem{Zhou2021Palceholder}
D.~Zhou, H.~Ye, and D.~Zhan, ``Learning placeholders for open-set recognition,'' in \emph{CVPR}, 2021, pp. 4401--4410.

\bibitem{Cevikalp2023From}
H.~Cevikalp, B.~Uzun, Y.~Salk, H.~Saribas, and O.~K{\"{o}}p{\"{u}}kl{\"{u}}, ``From anomaly detection to open set recognition: Bridging the gap,'' \emph{Pattern Recognition}, vol. 138, p. 109385, 2023.

\bibitem{sun2024overall}
L.~Sun and W.~Chu, ``Overall positive prototype for few-shot open-set recognition,'' \emph{Pattern Recognition}, vol. 151, p. 110400, 2024.

\bibitem{xia2023slcpl}
Z.~Xia, P.~Wang, G.~Dong, and H.~Liu, ``Spatial location constraint prototype loss for open set recognition,'' \emph{Computer Vision and Image Understanding}, vol. 229, p. 103651, 2023.

\bibitem{Lu2022PMALOS}
J.~Lu, Y.~Xu, H.~Li, Z.~Cheng, and Y.~Niu, ``{PMAL:} open set recognition via robust prototype mining,'' in \emph{AAAI}, 2022, pp. 1872--1880.

\bibitem{liu2023MGPL}
J.~Liu, J.~Tian, W.~Han, Z.~Qin, Y.~Fan, and J.~Shao, ``Learning multiple gaussian prototypes for open-set recognition,'' \emph{Information Sciences}, vol. 626, pp. 738--753, 2023.

\bibitem{wang2024exploring}
Y.~Wang, J.~Mu, P.~Zhu, and Q.~Hu, ``Exploring diverse representations for open set recognition,'' in \emph{AAAI}, 2024.

\bibitem{moon2022difficulty}
W.~Moon, J.~H. Park, H.~S. Seong, C.~Cho, and J.~Heo, ``Difficulty-aware simulator for open set recognition,'' in \emph{ECCV}, vol. 13685, 2022, pp. 365--381.

\bibitem{Neal2018OSCI}
L.~Neal, M.~L. Olson, X.~Z. Fern, W.~Wong, and F.~Li, ``Open set learning with counterfactual images,'' in \emph{ECCV}, vol. 11210, 2018, pp. 620--635.

\bibitem{wen2019comprehensive}
Y.~Wen, K.~Zhang, Z.~Li, and Y.~Qiao, ``A comprehensive study on center loss for deep face recognition,'' \emph{International Journal of Computer Vision}, vol. 127, pp. 668--683, 2019.

\bibitem{schroff2015facenet}
F.~Schroff, D.~Kalenichenko, and J.~Philbin, ``Facenet: A unified embedding for face recognition and clustering,'' in \emph{CVPR}, 2015, pp. 815--823.

\bibitem{OrtizCataln2013BioPatRecAM}
M.~Ortiz{-}Catalan, R.~Br{\aa}nemark, and B.~H{\aa}kansson, ``Biopatrec: {A} modular research platform for the control of artificial limbs based on pattern recognition algorithms,'' \emph{Source Code for Biology and Medicine}, vol.~8, p.~11, 2013.

\bibitem{DB4}
S.~Pizzolato, L.~Tagliapietra, M.~Cognolato, M.~Reggiani, H.~M{\"u}ller, and M.~Atzori, ``Comparison of six electromyography acquisition setups on hand movement classification tasks,'' \emph{PLoS ONE}, vol.~12, 2017.

\bibitem{DB7}
A.~Krasoulis, I.~Kyranou, M.~S. Erden, K.~Nazarpour, and S.~Vijayakumar, ``Improved prosthetic hand control with concurrent use of myoelectric and inertial measurements,'' \emph{Journal of NeuroEngineering and Rehabilitation}, vol.~14, 2017.

\bibitem{wang2024transformer}
Z.~Wang, J.~Yao, M.~Xu, M.~Jiang, and J.~Su, ``Transformer-based network with temporal depthwise convolutions for semg recognition,'' \emph{Pattern Recognition}, vol. 145, p. 109967, 2024.

\bibitem{zhang2022crossformer}
Y.~Zhang and J.~Yan, ``Crossformer: Transformer utilizing cross-dimension dependency for multivariate time series forecasting,'' in \emph{ICLR}, 2022.

\bibitem{millerclass}
D.~Miller, N.~Suenderhauf, M.~Milford, and F.~Dayoub, ``Class anchor clustering: A loss for distance-based open set recognition,'' in \emph{WACV}, 2021, pp. 3570--3578.

\bibitem{netzer2011svhn}
Y.~Netzer, T.~Wang, A.~Coates, A.~Bissacco, B.~Wu, A.~Y. Ng \emph{et~al.}, ``Reading digits in natural images with unsupervised feature learning,'' in \emph{NeurIPS workshop}, vol. 2011, no.~5, 2011, p.~7.

\bibitem{krizhevsky2009cifar}
A.~Krizhevsky \emph{et~al.}, ``Learning multiple layers of features from tiny images,'' 2009.

\bibitem{russakovsky2015imagenet}
O.~Russakovsky, J.~Deng, H.~Su, J.~Krause, S.~Satheesh, S.~Ma, Z.~Huang, A.~Karpathy, A.~Khosla, M.~Bernstein \emph{et~al.}, ``Imagenet large scale visual recognition challenge,'' \emph{International journal of computer vision}, vol. 115, pp. 211--252, 2015.

\bibitem{yoshihashi2019classification}
R.~Yoshihashi, W.~Shao, R.~Kawakami, S.~You, M.~Iida, and T.~Naemura, ``Classification-reconstruction learning for open-set recognition,'' in \emph{CVPR}, 2019, pp. 4016--4025.

\bibitem{liu2022orientational}
Z.~Liu, Y.~Fu, Q.~Pan, and Z.~Zhang, ``Orientational distribution learning with hierarchical spatial attention for open set recognition,'' \emph{IEEE Transactions on Pattern Analysis and Machine Intelligence}, vol.~45, no.~7, pp. 8757--8772, 2022.

\end{thebibliography}

\section{Supplementary material}
\subsection{Image Domain Verification}

To further verify the performance of our proposed approach, we make a comparative experiment on four public image datasets widely used for OSR performance evaluation, including MNIST, SVHN~\cite{netzer2011svhn}, CIFAR10~\cite{krizhevsky2009cifar} and TinyImageNet~\cite{russakovsky2015imagenet}. We compare our proposed method, PredIN, to the SOTA open-set recognition methods including Softmax, OpenMax~\cite{Bendale2016Openmax}, CROSR~\cite{yoshihashi2019classification}, CAC~\cite{millerclass}, PROSER~\cite{Zhou2021Palceholder}, CSSR~\cite{huang2022class}, CPN~\cite{Yang2020ConvolutionalPN}, ARPL~\cite{Chen2022ARPL}, ODL~\cite{liu2022orientational}, SLCPL~\cite{xia2023slcpl}, MGPL~\cite{liu2023MGPL}, m-OvR~\cite{park2024understanding} and DIAS~\cite{moon2022difficulty}. For a fair comparison, we use the same backbone as these methods. In terms of optimization, we use the SGD optimizer with a momentum value of $0.9$ and set the initial learning rate to $0.01$ which drops to $0.1$ at every 30 epochs. The parameters $\beta$, $\gamma$, $\alpha$, $m_1$, $m_2$ and feature dimension are set to $1.0$, $1.0$, $1.0$, $0.5$, $0.5$ and $128$, respectively. All results of SOTA methods are taken from the references except DIAS~\cite{moon2022difficulty}. As DIAS~\cite{moon2022difficulty} applies different dataset split ways, we reproduce their method using the recommended hyperparameters to unify the split information.
AUC performances are shown in \tablename~\ref{tab5}. Our approach achieves comparable performance to SOTA methods. This clearly demonstrates the general applicability of our approach across sEMG and image domains.

\begin{table}[!h]
    \centering
    \caption{Performance comparison (\%) with SOTAs in terms of AUC on four public image datasets. Results are averaged among five randomized splits. Best performances are highlighted in bold. * indicates the reproduced result to unify the split information.}
    \small
    \vspace{-0.1cm}
    \setlength{\tabcolsep}{0.1cm}{
        \begin{tabular}{lccccc}
        \toprule
        Methods & MNIST & SVHN & CIFAR10 & TinyIN\\
        \midrule
        Softmax  & 97.8 & 88.6 & 67.7 & 57.7   \\
        OpenMax (CVPR'16) & 98.1 & 89.4 & 69.5 & 57.6   \\
        CROSR (CVPR'19) & 99.1 & 89.9 & 88.3 & 58.9 \\
        CAC (WACV'21) & 99.1 & 94.1 & 80.1 & 76.0 \\
        PROSER (CVPR'21) & - & 94.3 & 89.1 & 69.3 \\
        CPN (TPAMI'22) & 99.0 & 92.6 & 82.8 & 63.9 \\
        ARPL (TPAMI'22)   & \textbf{99.6} & 96.3 & 90.1 & 76.2   \\
        ODL (TPAMI'22) & \textbf{99.6} & 95.4 & 88.5 & 74.6 \\
        SLCPL (CVIU'23) & 99.4 & 95.2 & 86.1 & 74.9 \\
        MGPL (Inf.Sci'23) & - & 95.7 & 84.0 & 73.0 \\
        m-OvR (Pattern Recognit.'24) & - & 95.7 & 89.5 & 75.3  \\
        DIAS* (ECCV'22)  & 99.5 & 94.7 & 90.3 & 76.8  \\
        \midrule  
        \textbf{PredIN} & \textbf{99.6} & \textbf{97.2} & \textbf{90.5} & \textbf{77.2}  \\
        \bottomrule
        \end{tabular}}
        \label{tab5}
        \vspace{-0.3cm}
\end{table}

\end{document}